\documentclass[10pt,twocolumn,letterpaper]{article}
\usepackage[T1]{fontenc} %

\usepackage[pagenumbers]{cvpr} %

\renewcommand{\paragraph}[1]{\vspace{.5em}\noindent\textbf{#1.}}

\newif\ifDEBUG
\DEBUGfalse

\newif\ifSUPPLEMENTAL
\SUPPLEMENTALtrue

\colorlet{cnncell}{Lavender!40!white}

\colorlet{xformercell}{Apricot!20!white}

\colorlet{ourcell}{RoyalBlue!20!white}

\ifDEBUG
    \newcommand{\NJE}[1]{\textcolor{red}{[NJE: #1]}}
    \newcommand{\PJ}[1]{\textcolor{blue}{[PJ: #1]}}
    \newcommand{\JD}[1]{\textcolor{purple}{[JD: #1]}}
    \newcommand{\YHL}[1]{\textcolor{olive}{[YHL: #1]}}
    \newcommand{\GKT}[1]{\textcolor{violet}{[GKT: #1]}}
    \newcommand{\BC}[1]{\textcolor{cyan}{[BC: #1]}}
\else
    \newcommand{\NJE}[1]{}
    \newcommand{\PJ}[1]{}
    \newcommand{\JD}[1]{}
    \newcommand{\YHL}[1]{}
    \newcommand{\GKT}[1]{}
    \newcommand{\GL}[1]{}
    \newcommand{\GR}[1]{}
    \newcommand{\BC}[1]{}
\fi

\DeclareRobustCommand{\blueuparrow}{\textcolor{blue}{$\uparrow$}}
\DeclareRobustCommand{\bluedownarrow}{\textcolor{blue}{$\downarrow$}}

\DeclareRobustCommand{\greencheckmark}{\textcolor{OliveGreen}{\checkmark}}

\usepackage{subcaption}
\usepackage{listings}
\usepackage{algorithm}
\usepackage{algpseudocode}
\usepackage{multirow}
\usepackage{placeins}

\NewDocumentCommand{\hlbox}{m}{%
  \begingroup
    \fboxsep=1.2pt
    \colorbox{blue!10}{\textbf{#1}}%
  \endgroup
}

\NewDocumentCommand{\hlboxeq}{m}{%
  \begingroup
    \fboxsep=1.2pt
    \colorbox{purple!10}{#1}%
  \endgroup
}

\newcommand{\algcomment}[1]{%
  \par\vspace{2pt}\noindent%
  {\raggedright\small\textit{#1}\par}%
}
\lstdefinestyle{mocov3}{
  backgroundcolor=\color{white},
  basicstyle=\fontsize{7pt}{7pt}\ttfamily\selectfont, %
  columns=fullflexible,
  breaklines=true,
  captionpos=b,
  commentstyle=\fontsize{7pt}{7pt}\color{ForestGreen},
  keywordstyle=\fontsize{7pt}{7pt}\color[rgb]{0.85,0.18,0.50},
  numbers=left,                          %
  numberstyle=\tiny\color{gray},         %
  numbersep=6pt,                         %
}

\definecolor{cvprblue}{rgb}{0.21,0.49,0.74}
\usepackage[pagebackref,breaklinks,colorlinks,allcolors=cvprblue]{hyperref}

\def\paperID{} %
\def\confName{CVPR}
\def\confYear{2026}

\newcommand{\mytitle}{}
\renewcommand{\mytitle}{AdaPerceiver: A Unified Framework for Adaptive Width, Depth, and Tokens}
\renewcommand{\mytitle}{AdaPerceiver: Transformers with Adaptive Width, Depth, and Tokens}
\renewcommand{\mytitle}{AdaPerceiver: Efficient Transformers by Adapting Width, Depth, and Tokens}
\renewcommand{\mytitle}{AdaPerceiver: Transformers with Adaptive Width, Depth, and Tokens}

\title{\mytitle}

\author{Purvish Jajal\\
Purdue University\\
West Lafayette, IN, USA\\
{\tt\small pjajal@purdue.edu}
\and
Nick John Eliopoulos \\
Purdue University\\
West Lafayette, IN, USA\\
{\tt\small neliopou@purdue.edu}
\and
Benjamin Shiue-Hal Chou\\
Purdue University\\
West Lafayette, IN, USA\\
{\tt\small chou150@purdue.edu}
\and
George K. Thiruvathukal\\
Loyola University Chicago\\
Chicago, IL, USA\\
{\tt\small gkt@cs.luc.edu}
\and
Yung-Hsiang Lu\\
Purdue University\\
West Lafayette, IN, USA\\
{\tt\small yunglu@purdue.edu}
\and
James C. Davis\\
Purdue University\\
West Lafayette, IN, USA\\
{\tt\small davisjam@purdue.edu}
}

\begin{document}
\maketitle

\begin{abstract}
Modern transformer architectures achieve remarkable performance across tasks and domains but remain rigid in how they allocate computation at inference time.
Real-world deployment often requires models to adapt to diverse hardware and latency constraints, yet most approaches to dynamic computation focus on a single axis --- such as reducing the number of tokens.
We present a novel capability: AdaPerceiver, the first transformer architecture with \textit{unified adaptivity across depth, width, and tokens within a single model}.
We propose an architecture that supports adaptivity along these axes.
We couple this with an efficient joint training regime that ensures the model maintains performance across its various configurations.
We evaluate AdaPerceiver on image classification, semantic segmentation, and depth estimation tasks.
On image classification, AdaPerceiver expands the accuracy-throughput Pareto front.
It achieves 85.4\% accuracy while yielding 36\% higher throughput than FlexiViT-L.
On dense prediction, AdaPerceiver matches ViT-H/14 while having $\sim$26x fewer encoder FLOPs (floating-point operations) on semantic segmentation and depth estimation.
Finally, we show how AdaPerceiver equipped with a policy can maintain ImageNet1K accuracy ($\pm0.1$ percentage points) while reducing FLOPs by $24-33\%$.
\end{abstract}

\section{Introduction}
\label{sec:intro}
Adaptivity---the ability to flexibly allocate computation based on input complexity or resource constraints---enables efficient machine learning systems~\cite{devvrit2024matformer, han2021dynamic, haberer2024hydravit}.
For transformer models, adaptivity can be applied along three primary axes: 
\emph{tokens} (number of tokens processed),
\emph{depth} (number of layers executed),
and \emph{width} (embedding dimension). 
Each axis offers a different trade-off: 
  tokens improves dense prediction performance but increase computational costs quadratically~\cite{wang2025scaling};
  depth governs representational refinement but increases computational costs linearly~\cite{jastrzkebski2017residual};  
  and
  width enhances capacity but affects computational costs of feed-forward networks~(FFNs)~\cite{devvrit2024matformer}.

Jointly supporting adaptivity across all three axes is combinatorially challenging. %
In practice, common modern architectures such as Vision Transformers (ViTs)~\cite{dosovitskiy2020image} operate instead with a fixed computational budget.
Every input is processed with the same number of layers, tokens, and parameters.
Although adaptive models have been introduced~\cite{haberer2024hydravit,beyer2023flexivit,devvrit2024matformer, rao2021dynamicvit, devvrit2024matformer}, they typically restrict adaptivity along one or two axes, failing to capture the full range of trade-offs in modern networks (\cref{tab:adaptive_axes}). 
No unified framework captures all three axes within a single model.

\begin{table}[ht]
\centering
\caption{Comparison of adaptive dimensions across models.
}
\setlength{\tabcolsep}{7.5pt}
\begin{tabular}{lccc}
\toprule
\textbf{Model} & \textbf{Tokens} & \textbf{Depth} & \textbf{Width} \\
\midrule
FlexiViT~\cite{beyer2023flexivit}     & \greencheckmark & -- & -- \\
MatFormer~\cite{devvrit2024matformer}    & -- & -- & \greencheckmark \\
HydraViT~\cite{haberer2024hydravit}     & -- & -- & \greencheckmark \\
DynaBERT~\cite{hou2020dynabert}    & -- & \greencheckmark & \greencheckmark \\
\midrule
\textbf{AdaPerceiver} (Ours) & \greencheckmark & \greencheckmark & \greencheckmark \\
\bottomrule
\end{tabular}
\label{tab:adaptive_axes}
\end{table}

We address this gap with the \textbf{Ada}ptive \textbf{Perceiver}, a novel architecture that unifies adaptivity across all three axes---tokens, width, and depth---within a single model configurable at inference time.
We show how to create and train a single architecture with adaptivity across each axis.
\textbf{AdaPerceiver} combines block-masked attention for token adaptivity, Matryoshka FFNs for width adaptivity, and early-exiting for depth adaptivity.
We then show how to train this architecture without the combinatorial complexity of joint training (\cref{eq:meth:training:joint}) nor the noisier configuration sampling~\cite{haberer2024hydravit}.
Our \textit{once-for-all training strategy} allows for the \textit{learning of adaptivity across all axes in a single forward pass}.
The result is a single model that can be configured at inference-time, with favourable accuracy-efficiency trade-offs. %

We evaluate AdaPerceiver on three vision tasks: image classification, semantic segmentation, and depth estimation.
For ImageNet1K classification, AdaPerceiver expands the accuracy-throughput Pareto frontier, achieving 85.4\% accuracy while yielding 36\% higher throughput than FlexiViT-L.
On ADE20K semantic segmentation, we achieves comparable mIOU to ViT-H, and on NYUv2 depth estimation we outperform ViT-H with $\sim$26$\times$ fewer FLOPs.
Finally, we show that AdaPerceiver, when configured with a suitable policy, can maintain ImageNet1K accuracy ($\pm0.1$~percentage points) while reducing FLOPs by $24-33\%$.

\underline{In sum, our contributions are}:
\begin{itemize}[noitemsep]
    \item We propose \textbf{AdaPerceiver}, an adaptive architecture that enables compute–accuracy trade-offs along three axes: tokens, depth, and width within a single model.
    AdaPerceiver can dynamically adapt its computational footprint at inference time to meet diverse constraints, from resource-limited devices to high-accuracy settings. 
    \item We develop a \textbf{once-for-all training recipe} that leverages structured masking to jointly train multiple sub-networks within a \textit{single forward pass}, ensuring robust performance across all dimensions of adaptivity.
\end{itemize}

\section{Related Work}
\label{sec:rw}

Our approach builds on two lines of prior research:
  \textit{adaptive models}
    (\cref{sec:rw:adaptive-models})
  and
  the \textit{Perceiver architecture} (\cref{sec:rw:perceiver}).

\subsection{Adaptive Models}
\label{sec:rw:adaptive-models}

Adaptivity in deep learning has been explored through two main traditions: dynamic (\ie conditional) neural networks (NNs)~\cite{bengio2015conditional, han2021dynamic}, and elastic models~\cite{cai2019once, devvrit2024matformer}.

\paragraph{Dynamic Neural Networks}
Dynamic neural networks adapt computation on a per-input basis, allocating compute or parameters depending on the difficulty or content of the input.
Approaches include early-exiting strategies~\cite{teerapittayanon2016branchynet, wang2018skipnet, wolczyk2021zero, zhou2020bert, raposo2024mixture} and pruning techniques~\cite{gao2018dynamic, xu2022evo, rao2021dynamicvit, bolya2022token, yin2022vit}.

\paragraph{Elastic Models}
In contrast, the elastic model tradition focuses on training a \textit{single model} that can be executed at multiple capacities under user-defined compute budgets~\cite{devvrit2024matformer, valipour2023sortednet, haberer2024hydravit, cai2024flextron, hou2020dynabert}. 
Early work such as Once-for-All networks~\cite{cai2019once} demonstrated that convolutional networks can be trained to support a set of sub-networks that trade accuracy for efficiency at inference time. 
Subsequent work extends this idea to Transformer architectures, enabling flexible inference across 1--2 dimensions: tokens, depths, or widths. 
Width-adaptive models such as MatFormer, HydraViT, and Flextron~\cite{devvrit2024matformer, valipour2023sortednet, haberer2024hydravit, cai2024flextron} train shared-weight sub-networks that operate at varying hidden dimensions, while DynaBERT and SortedNet~\cite{hou2020dynabert, valipour2023sortednet} explore joint width–depth adaptivity. 
Token-adaptivity has been studied in FlexiViT~\cite{beyer2023flexivit}, which supports varying patch sizes at inference---and thus token counts---within a single model.

Existing training strategies for these models are either costly (relying on multiple forward-passes per configuration~\cite{devvrit2024matformer}) or noisy (stochastic training approaches that sample configurations~\cite{haberer2024hydravit, valipour2023sortednet, cai2024flextron, beyer2023flexivit}).

\paragraph{Comparison to Our Work}
AdaPerceiver combines elements of both traditions.  
Like dynamic neural networks, it supports per-input adaptivity: configurations can be selected at runtime, \eg by a learned policy (see \cref{sec:eval:policies}).  
Similar to elastic models, we train a single shared-weight model to support flexible configurations.  
However, unlike prior elastic models, AdaPerceiver supports simultaneous adaptivity across token, depth, and width axes.  
For our novel training approach, %
we structure the network such that multiple configurations can be \textit{jointly optimized within a single forward pass}.
Training AdaPerceiver does not require multiple forward evaluations, with less reliance on stochastic configuration sampling.

\subsection{Perceiver Architecture}
\label{sec:rw:perceiver}
Perceiver architectures follow an \textit{encode-process-decode} paradigm:
  inputs are \textit{encoded} via attention into a fixed set of latent tokens (the latent stream);
  this latent stream is \textit{processed} through iterative transformer layers;
  and finally
  \textit{decoded} to produce outputs. 
The original Perceiver introduced a fixed-size latent stream that decoupled input size from internal computation, enabling scalability to large and multi-modal data~\cite{perceiverjaegle21a}.
PerceiverIO extended this idea by introducing an output query mechanism, allowing latent representations to be decoded into arbitrarily sized outputs~\cite{jaegle2021perceiverio}. 
Subsequent variants further developed this direction. PerceiverAR~\cite{hawthorne2022percevierar} adapted the architecture for autoregressive modeling, while the Hierarchical Perceiver (HiP)~\cite{carreira2022hip} incorporated locality and hierarchical structure to improve efficiency while maintaining generality.

\begin{figure*}[t]
    \centering
    \includegraphics[width=0.80\linewidth,trim=27 10 30 25,clip]{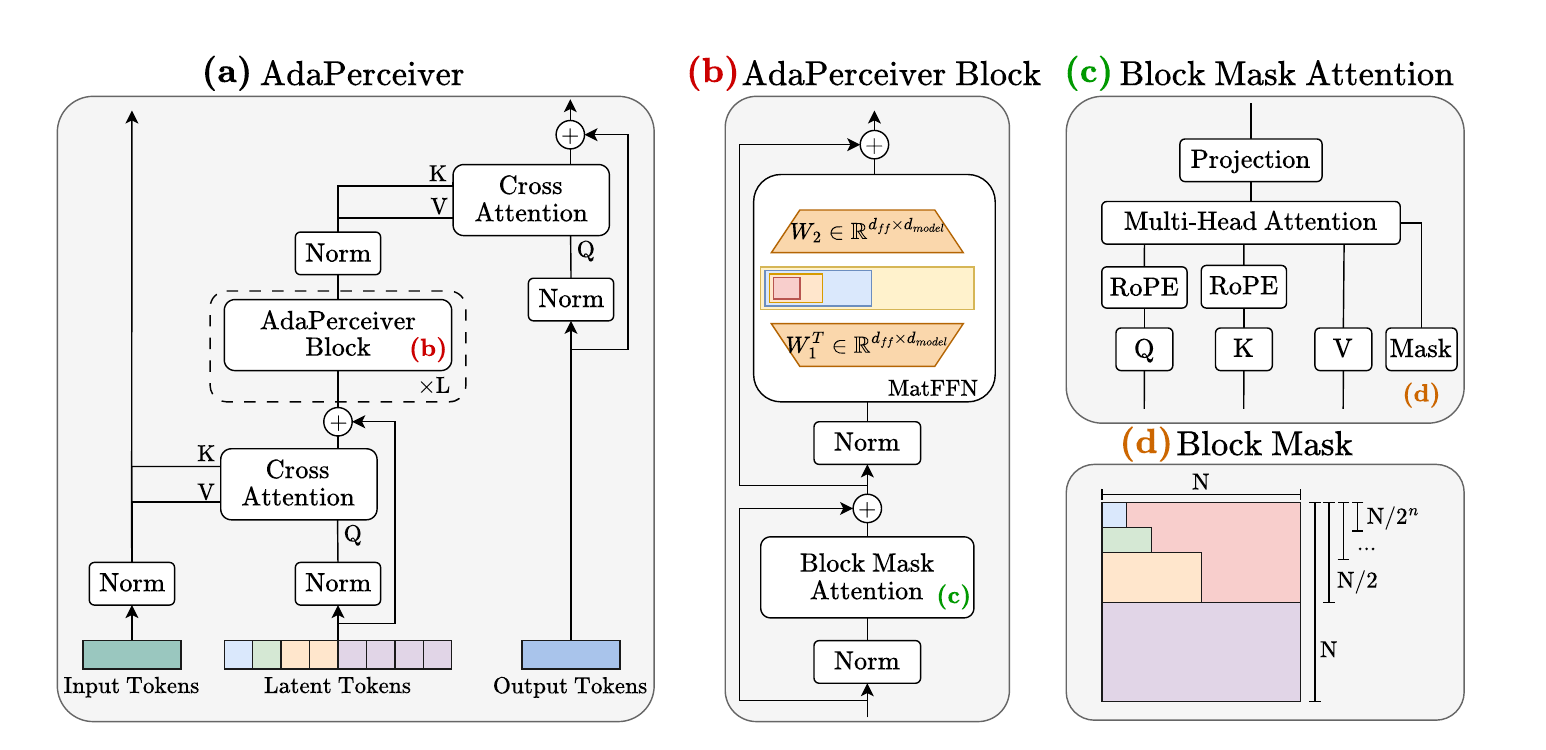}
    \caption{
    Overview of Adaptive Perceiver (AdaPerceiver). 
    \textbf{(a)} AdaPerceiver architecture.
    The AdaPerceiver architecture consists of three streams: input, output and latent. 
    Cross-attention blocks map input tokens to latent tokens and read out latent tokens to output tokens.
    The latent stream allows for an \textit{adaptive embedding} and \textit{adaptive token dimensions}.
    \textbf{(b)} The AdaPerceiver block follows a standard pre-norm transformer architecture~\cite{dosovitskiy2020image}, but replaces bi-directional self-attention with \textit{block mask attention} (\textit{c}). 
    Its feed-forward network (FFN) is similar to MatFormer~\cite{devvrit2024matformer}, enabling \textit{adaptive embedding dimensions}.
    \textbf{(c)} \textit{Block mask attention}, is akin to self-attention in ViTs~\cite{dosovitskiy2020image} but instead applies Rotary Positional Encoding (RoPE) on the Q and K matrices~\cite{su2024roformer} and masks attention maps as shown in \textit{(d)}. 
    This design enables \textit{adaptive token dimensions}.
    \textbf{(d)} Visualization of block masking for $N$ tokens: \textit{Red} denotes masked tokens, while other \textit{colours} indicate unmasked tokens.
    Masking restricts attention interactions at each latent token granularity, ensuring that later tokens can attend to earlier ones, but not vice versa. 
    We elaborate in \cref{sec:method}.
    \textit{N.B.} The $\log_2$-spaced token granularity is arbitrary.
    }
    \label{fig:method:architecture}
\end{figure*}

\paragraph{Comparison to Our Work}
Prior work on the Perceiver family studies generality and scalability across modalities.
Their latent processing streams are fixed once trained. 
We introduce adaptivity into this latent stream, enabling control over the amount of computation allocated to each input.

\section{Adaptive Perceiver}
\label{sec:method}
In this section we describe \textit{AdaPerceiver}, an adaptive transformer architecture that enables adaptivity along three axes: token, depth, and width.
\cref{sec:meth:requirements} outlines the requirements for adaptivity,
\cref{sec:meth:arch} introduces the AdaPerceiver architecture and describes how it meets the requirements,
and \cref{sec:meth:training} details our training procedure.

\subsection{Requirements and Challenges}
\label{sec:meth:requirements}
Following prior work~\cite{devvrit2024matformer, haberer2024hydravit}, learning an adaptive network requires: 
(a) \textit{architectural support} for controllable token count, depth, and width; and
(b) a \textit{training scheme} that allows for efficient training across the adaptive axes.

For (a), configurability allows the model to vary its computational cost and representational capacity, allowing efficiency/accuracy trade-offs.
In practice, exposing such configurability is straightforward at the implementation level, \eg tensor slicing. 
The challenge lies in training the induced sub-networks (b).
Learning across the sub-networks must be done efficiently. 
Training each sub-network independently is infeasible; prior work~\cite{devvrit2024matformer, cai2019once, haberer2024hydravit, valipour2023sortednet} therefore uses joint optimization across sub-networks. 
However, as adaptivity spans multiple axes, the number of sub-networks grows combinatorially, motivating a co-design of architecture and training to maintain efficiency.

\subsection{Architecture}
\label{sec:meth:arch}
AdaPerceiver exposes adaptivity while enabling single-pass joint optimization across configurations.
Computation is structured so one encoder forward pass yields features supporting many sub-networks, avoiding multiple passes~\cite{devvrit2024matformer} or high-variance configuration sampling~\cite{haberer2024hydravit}.

\subsubsection{Overview}
\label{sec:meth:arch:over}
AdaPerceiver extends the PerceiverIO architecture~\cite{jaegle2021perceiverio} by introducing adaptivity in depth, width, and tokens. 
PerceiverIO decouples the number of input and output tokens from the latent tokens, enabling token adaptivity since the number of latents can vary independently of the input and output. 
This separation allows structure to be imposed on the latent tokens, making it possible to jointly optimize multiple token configurations within a single forward pass (see \cref{sec:appendix:flexi} for why other token-adaptive methods, such as FlexiViT~\cite{beyer2023flexivit}, do not allow this). 

As illustrated in \cref{fig:method:architecture}~(a), AdaPerceiver consists of three interacting streams: \textit{input}, \textit{latent}, and \textit{output}.
Cross-attention layers map input tokens to latent tokens and decode the latents to output tokens. 
Within the latent stream, a series of \textit{AdaPerceiver Blocks} iteratively refine the latent representation.
\textit{Depth adaptivity} is realized through early exiting within the latent stream;
\textit{Width adaptivity} through Matryoshka feed-forward layers~\cite{devvrit2024matformer}; 
and
\textit{token adaptivity} by varying the number of latent tokens.
For further details see~\cref{sec:appendix:architecture}.

\subsubsection{Latent Tokens}
\label{sec:meth:arch:latent}
To support token adaptivity we learn a single latent vector that is broadcast to the desired number of latent tokens.
To distinguish the broadcasted latent tokens we rely upon 1D Rotary Positional Embedding (RoPE)~\cite{su2024roformer} applied within the attention mechanism. 

\paragraph{Rationale} 
This choice offers two advantages. 
First, 1D RoPE does not tie the latent representation to the spatial structure of the input, allowing the latent sequence to serve as an abstract, modality-agnostic processing space. 
Second, because RoPE provides relative positional encoding, it naturally supports \textit{extrapolation} beyond the training token length, enabling the model to process arbitrary numbers of latent tokens (see \cref{fig:eval:qual:tokens}).
Overall, this design enables the latent sequence to be a flexible, supporting variable-length configurations (\textit{cf.} \cref{sec:appendix:latent,fig:appendix:image-class:inter-extrap,fig:appendix:seg:inter-extrap,fig:appendix:depth-est:inter-extrap}).

\paragraph{Alternatives}
In lieu of a single latent token, a fixed latent array can be learned following PerceiverIO~\cite{jaegle2021perceiverio}.
However, this increases the number of parameters and makes extrapolation beyond training length non-trivial (see \cref{sec:appendix:latent}).
As such, we do not consider it.

\subsubsection{AdaPerceiver Block}
\label{sec:meth:arch:block}
Each AdaPerceiver Block (\cref{fig:method:architecture}~(b)) follows a standard pre-norm transformer design but replaces bidirectional self-attention with \textit{block mask attention} and the feed-forward network with a Matryoshka variant (see \cref{alg:appendix:matlinear} in \cref{sec:appendix:pseudo} for a concise implementation). 

\subsubsection{Block Mask Attention}
\label{sec:meth:arch:mask-attn}
Block mask attention (\cref{fig:method:architecture}~(c)) follows ViT-style self-attention~\cite{dosovitskiy2020image}, but applies 1D RoPE to the query and key matrices and introduces a structured attention mask (\cref{fig:method:architecture}~(d)). 
Block masking constrains attention within token groups, such that later token groups can attend to earlier ones but not vice versa. 

\paragraph{Rationale}
This design allows us to train \textit{as if} the model has seen different sequence lengths in a single forward pass, parallelizing the token adaptivity training.
For our intuition, see \cref{sec:appendix:block-mask}.

\paragraph{Alternatives}
Fully bidirectional attention pattern as is common in ViTs~\cite{dosovitskiy2020image} prevents parallelization across token granularities, requiring separate passes for each token configuration.
We attempted this in our early experiments but found slow, noisy convergence. %
Nevertheless, we find that bidirectional attention can often be enabled at inference without performance degradation; see Appendix~\cref{fig:appendix:image-class:pareto-bidir-full}.

\subsection{Training}
\label{sec:meth:training}
\begin{algorithm}[t]
\caption{
AdaPerceiver Training. See \cref{alg:appendix:adaperceiver} for commented version.}
\label{alg:adaperceiver}
\lstset{style=mocov3}
\begin{lstlisting}[language=python,escapechar=@]
width_choices = [...] # set of width configurations
token_grans = [...] # set of token granularities
mask = create_block_mask(latent_token_grans)

class AdaPerceiver(...):
  def forward(x, mask, widths):
    # N = maximum latents tokens used during training.
    # M = # of output tokens
    z = ... # [B, N]
    o = ... # [B, M] # output tokens

    @\hlboxeq{(\cref{eq:meth:training:encoder})}@ 
    # Encode input to latents (Q=latents, K/V=x)
    latents = cross_attention(sink=latents, src=x)
    # Process latents with AdaPerceiver blocks
    z_L, z_ls = forward_blocks(latents, mask, widths)

    @\hlbox{Token Loss, \cref{sec:meth:training:token}}@
    # Decode outputs at token granularities @\hlboxeq{(\cref{eq:meth:training:token:ca})}@
    outputs, inter_outputs = [], []
    for t in token_grans:
      o_t = cross_attention(sink=o, src=z_L[:, :t])
      outputs.append(o_t)
    @\hlbox{Depth Loss, \cref{sec:meth:training:depth}}@
    # Decode outputs at multiple depths @\hlboxeq{(\cref{eq:meth:training:depth:ca})}@ 
    for z_l in z_ls:
      t = sample(latent_grans)
      o_l = cross_attention(sink=o, src=z_l[:, :g])
      inter_outputs.append(o_l)
    return outputs, inter_outputs

model = AdaPerceiver(...)
for x, y in dataloader:
  @\hlbox{Width Loss (implicit), \cref{sec:meth:training:width}}@ 
  # sample width per sample in batch (B = batch size)
  widths = [sample(width_choices) for _ in range(B)]
  outputs, inter_outputs = model(x, mask, widths)
  @ \hlboxeq{Uses (\cref{eq:meth:training:joint_ada,eq:meth:training:token,eq:meth:training:depth})}@
  loss = loss_fn(outputs, y) + loss_fn(inter_outputs, y)
  loss.backward()
  ...
\end{lstlisting}
\end{algorithm}

\paragraph{Notation}
We denote by $N$ the maximum number of latent tokens, $M$ the number of output tokens, 
$L$ the number of latent blocks (depth), and $d$ the embedding dimension. 
We define three configuration sets: 
$\mathcal{T}$ for token granularities, 
$\mathcal{W}$ for width configurations, and 
$\mathcal{D}$ for depths. 
Each adaptive sub-network is indexed by a tuple $(t, w, l)$ where 
$t \in \mathcal{T}$, $w \in \mathcal{W}$, and $l \in \mathcal{D}$.

\paragraph{Training Objective}
We jointly optimize over the sub-networks induced by the adaptive axes.
For a batch $B$,
\begin{equation}
    \mathcal{L}_{\text{joint}} =
    \frac{1}{B}\sum_{i=1}^{B}
    \sum_{t \in \mathcal{T}}\sum_{w \in \mathcal{W}}\sum_{l \in \mathcal{D}}
    \mathcal{L}\!\big(f_{(t, w, l)}(x_i),\, y_i\big),
    \label{eq:meth:training:joint}
\end{equation}
where $\mathcal{L}$ is the loss function, $y_i$ is the target label, and 
$f_{(t, w, l)}$ denotes the model $f$ instantiated with configuration $(t, w, l)$.
Naively evaluating this objective requires a separate forward pass for every configuration,
incurring $\mathcal{O}(|\mathcal{T}|\cdot|\mathcal{W}|\cdot|\mathcal{D}|)$ cost~\cite{devvrit2024matformer, haberer2024hydravit}.
Although stochastic sampling strategies~\cite{haberer2024hydravit} reduce this cost, they converged slowly in our early experiments and thus we did not pursue them.

Instead, the AdaPerceiver architecture enables joint optimization within a \textit{single encoder forward pass}, 
with $O(|\mathcal{T}| \cdot |\mathcal{D}|)$ additional passes through the (lightweight) output cross-attention.
This is tractable because output cross-attention constitutes $\approx2\%$ of total parameters.
We decompose the joint training objective as:
\begin{equation}
    \mathcal{L}_{\text{joint}} =
    \frac{1}{B}\sum_{i=1}^{B}
    \Big[\mathcal{L}_{\text{token}}(x_i, y_i, w_i)
        + \mathcal{L}_{\text{depth}}(x_i, y_i, w_i)\Big],
    \label{eq:meth:training:joint_ada}
\end{equation}
where $w_i\sim\mathrm{Uniform}(\mathcal{W})$ is a sampled width for each example.
For a given input $x_i$, the encoder produces intermediate latent representations:
\begin{equation}
    \{z_l\}_{l\in\mathcal{D}} = \mathtt{Encoder}(x_i; w_i),
    \label{eq:meth:training:encoder}
\end{equation}
with $z_l \in \mathbb{R}^{N\times d}$ denoting the latent tokens after the $l$-th block.
The encoder is evaluated once per sampled width.
The resulting $\{z_l\}$ is reused to compute token- and depth-level losses through lightweight cross-attention readouts.

\cref{alg:adaperceiver} shows the full training procedure.

\subsubsection{Token Loss}
\label{sec:meth:training:token}
To train for token adaptivity, we leverage block-mask attention (\cref{sec:meth:arch}) and the final cross-attention readout to simulate multiple token granularities in a single forward pass. 
Given the encoder outputs $\{z_l\}$, we compute the token loss using the final latent representation $z_{L}\in\mathbb{R}^{N\times d}$ and output tokens $o\in\mathbb{R}^{M\times d}$ as:
\begin{align}
    o_t &= \mathtt{CrossAttention}(o,\; z_{L}[:t]), \label{eq:meth:training:token:ca}\\
    \mathcal{L}_{\text{token}}(x_i, y_i, w_i) &= \sum_{t\in \mathcal{T}} \mathcal{L}(o_t, y_i).
    \label{eq:meth:training:token}
\end{align}
That is, after a single forward pass, we slice the first $t$ latent tokens from $z_{L}$, read out each token granularity $t$ via cross-attention, and compute their respective losses.

\subsubsection{Depth Loss}
\label{sec:meth:training:depth}
To train for depth adaptivity, we supervise intermediate representations at multiple depths~\cite{teerapittayanon2016branchynet, jiang2024tracing}.
For each depth $l\in\mathcal{D}$, we sample a token granularity $t_l\sim\mathrm{Uniform}(\mathcal{T})$, \ie uniformly from the set $\mathcal{T}$, and compute the
readout:
\begin{align}
    o_l &= \mathtt{CrossAttention}(o,\; z_l[:t_l]), \label{eq:meth:training:depth:ca}\\
    \mathcal{L}_{\text{depth}}(x_i, y_i, w_i) &= \sum_{l\in \mathcal{D}} \mathcal{L}(o_l, y_i).
    \label{eq:meth:training:depth}
\end{align}
Thus, the 
  encoder latents $\{z_l\}$ are reused across depths; only the readouts differ by sampled token granularity.
\subsubsection{Width Loss}
\label{sec:meth:training:width}
Width adaptivity is trained implicitly by sampling a width configuration $w_i\sim\mathrm{Uniform}(\mathcal{W})$ for each example.
Because width affects the encoder forward pass itself, its gradients propagate through both $\mathcal{L}_{\text{token}}$ and $\mathcal{L}_{\text{depth}}$, making a separate width loss unnecessary.
Per-batch width sampling is also viable, but we observed slower convergence.

\section{Evaluation}
\label{sec:evaluation}
We evaluate AdaPerceiver quantitatively and qualitatively, and provide an example of how it supports per-input adaptivity.
\cref{sec:eval:experimental} describes our experimental setup.
\cref{sec:eval:classification} evaluates AdaPerceiver on ImageNet-1K classification. 
\cref{sec:eval:dense} evaluates AdaPerceiver on dense-prediction tasks.
\cref{sec:eval:qual} qualitatively analyzes AdaPerceiver's learned representations.
Finally, \cref{sec:eval:policies} demonstrates how AdaPerceiver can be augmented with a policy to handle per-input adaptivity.

\subsection{Experimental Setup}
\label{sec:eval:experimental}
This section details the model configuration, training procedure, datasets, and evaluation protocols. 
We train on 16~NVIDIA H100 SXM 80GB GPUs.
We evaluate models on a NVIDIA~A100 80GB (PCIe) GPU.

\paragraph{Model and Adaptivity Configuration}
We base our model on shape-optimized ViT architectures~\cite{alabdulmohsin2023getting}, specifically the public implementations of SoViT-150M in the \texttt{timm} library~\cite{rw2019timm}.
Thus our model is 21 layers, with an embedding dimension of $832$.
We choose the following configuration: $\mathcal{T}=\{32, 64, 96, 128,1 92, 256\}$, 
$\mathcal{D}=\{1,2,\ldots,21\}$
, and $\mathcal{W}=\{416, 624, 832\}$.
Further details are in \cref{sec:appendix:architecture}.

\paragraph{Training Procedure}
We pre-train our model on ImageNet-12K~\cite{timm_imagenet12k_wds} for 150 epochs.
We follow the general training procedure from~\cref{sec:meth:training}, with a few practical modifications.
To reduce overall training time, we use logit distillation from a larger pre-trained Vision Transformer (ViT-H) into Adaptive Perceiver throughout training.
We use a curriculum~\cite{bengio2009curriculum, hacohen2019power} to learn adaptivity:
  we first train the model to adapt over the \textit{token} dimension,
  then jointly over \textit{token} and \textit{depth}, and finally over all three dimensions. 
For dense prediction tasks, we then add feature distillation from the teacher.
Further task-specific training details are in the respective sections.
For full information, see~\cref{sec:appendix:training}. 

\subsection{Image Classification}
\label{sec:eval:classification}
We evaluate AdaPerceiver on the ImageNet-1K classification benchmark.
We compare AdaPerceiver to publicly available elastic architectures: 
MatFormer (MatViT)~\cite{devvrit2024matformer}, FlexiViT~\cite{beyer2023flexivit}, and HydraViT~\cite{haberer2024hydravit}.
For completeness,~\cref{tab:vit-benchmark} has results on standard non-elastic baselines.

\paragraph{Training}
For the AdaPerceiver, we freeze our pre-trained backbone and fine-tune only the linear classification head, the output tokens, and the output cross-attention module responsible for decoding the latent representations.
All compared models are evaluated using their publicly released pre-trained weights, and no further fine-tuning is applied.

\begin{figure}[!tb]
    \centering
    \includegraphics[width=1\linewidth]{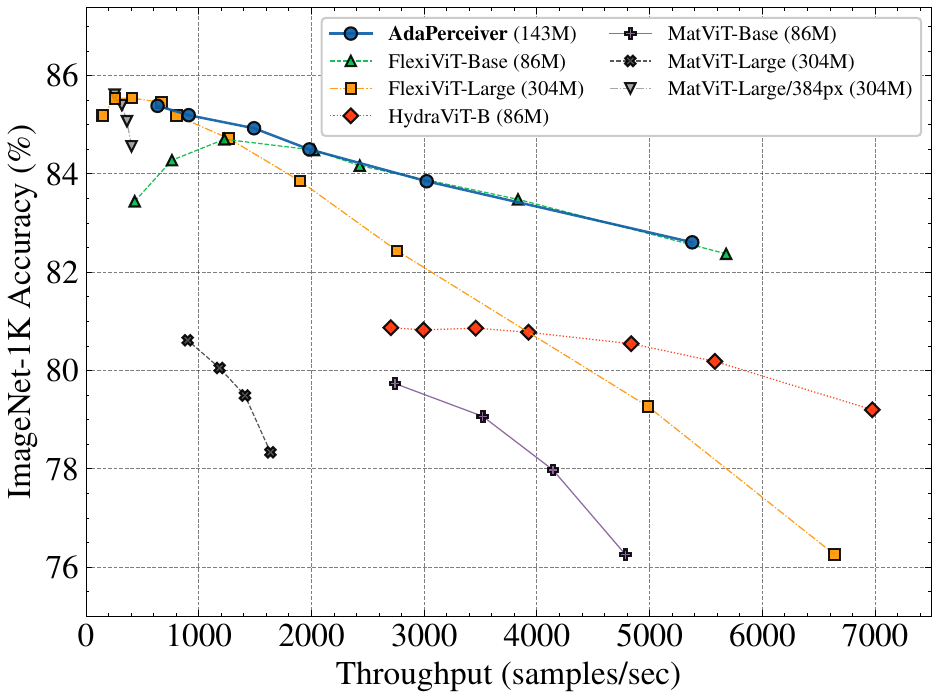 }
    \caption{
    \textbf{ImageNet-1K Evaluation.} 
    Accuracy vs. throughput (samples/sec) comparison of AdaPerceiver against state-of-the-art adaptive architectures.
    Each point corresponds to a distinct configuration. 
    AdaPerceiver's width ($w=832$) and depth ($l=21$) are fixed while varying the number of tokens.
    It achieves the best accuracy–efficiency trade-off: in the high-accuracy regime it matches large models, and in the high-throughput regime it matches FlexiViT-Base.
    Throughput is measured with batch size 512.
    This figure is a truncated version of Appendix~\cref{fig:appendix:image-class:pareto-bidir-merged}.
    }
    \label{fig:eval:image-class:pareto}
\end{figure}

\paragraph{Results}
\cref{fig:eval:image-class:pareto} depicts AdaPerceiver's results on ImageNet-1K, compared with other adaptive architectures (\textit{cf.} Appendix~\cref{fig:appendix:image-class:pareto-bidir-merged} and \cref{tab:vit-benchmark} for full data).
AdaPerceiver expands the Pareto frontier of accuracy-throughput tradeoffs.
In the high-accuracy regime, AdaPerceiver achieves 85.4\% accuracy, which is $0.1$ lower than FlexiViT-L, but with $36\%$ higher throughput.
In the high-throughput regime, AdaPerceiver (5378 img/s) nearly matches FlexiViT-B (5676 img/s), while achieving $0.2$ percentage points higher accuracy.
Meanwhile, the nearest FlexiViT-L configuration achieves 4970 img/s but has $3.3$ percentage points \emph{lower accuracy}.%

Between these two extremes, AdaPerceiver outcompetes FlexiViT-B and FlexiViT-L, achieving a Pareto-optimal tradeoff.
These results demonstrate that AdaPerceiver, a single model, can interpolate between high-accuracy and high-throughput at runtime \emph{while being Pareto-optimal}.

\begin{figure}[!tb]
    \centering\includegraphics[width=1\linewidth]{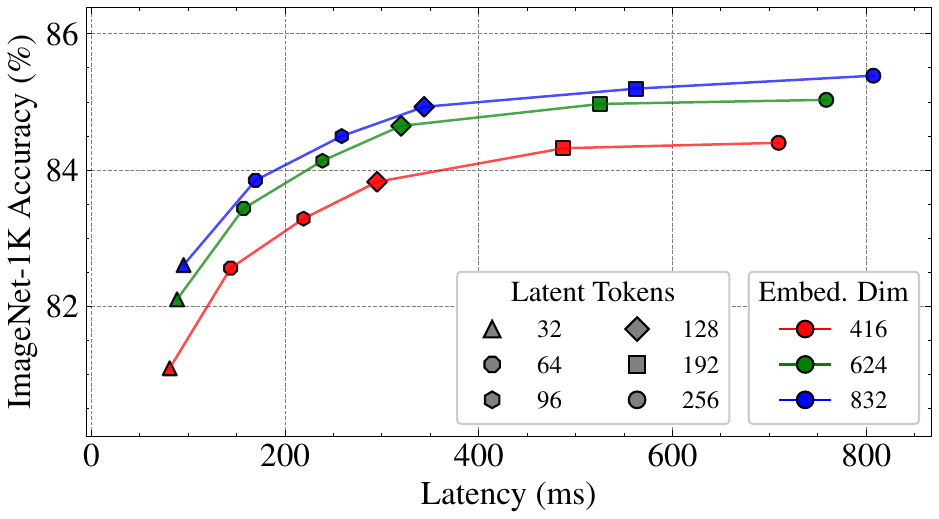}
    \caption{
    \textbf{ImageNet-1K Configuration Tradeoffs.} 
    Accuracy vs. latency (ms) for AdaPerceiver under varying embedding dimensions and numbers of latent tokens.
    Note: \textit{each configuration (point) does not require retraining}.
    Increasing the embedding dimension improves accuracy,
    while reducing the number of latent tokens decreases latency.
    }
    \label{fig:eval:image-class:token-matdim-tradeoff}
\end{figure}

\cref{fig:eval:image-class:token-matdim-tradeoff} depicts trade-offs between token-width configurations (fixed-depth). 
Reducing tokens has lower impact on accuracy than reducing embedding dimension.
This pattern holds across token-depth trade-offs, as shown in Appendix \cref{fig:appendix:image-class:token-depth}, increasing depth monotonically improves accuracy, and reducing tokens has a smaller impact than reducing width.


\subsection{Dense Prediction}
\label{sec:eval:dense}
To understand how adaptivity affects dense prediction, we evaluate on semantic segmentation and depth estimation tasks.
Because our intention is characterization of adaptivity rather than state-of-the-art performance, we follow the simple dense prediction protocols from \cite{tips_paper}. 
For each task, we compare AdaPerceiver against its teacher model (from distillation) and smaller variants of its teacher.
We compare against variants of the teacher which are not distilled.

\subsubsection{Semantic Segmentation}
\label{sec:eval:dense:segmentation}
We evaluate on the ADE20K dataset~\cite{zhou2017scene}.

\paragraph{Training}
We use the \textit{linear} head setup from \cite{tips_paper, oquab_dinov2_2023}.
For both AdaPerceiver and the baseline comparison, we attach a linear layer to the network, upsample the logit predictions to the input resolution, and apply a cross-entropy loss.
For AdaPerceiver, we retain the multi-layer perceptron (MLP) adapter
 from feature distillation and attach the linear head.

\begin{table}[t]
    \centering
    \caption{\textbf{ADE20K Evaluation.} Mean IoU and Forward GFLOPs (encoder only).
    For AdaPerceiver, the number of \textit{output tokens is 1369} (matching ViT-L and -H).
    We vary the number of  \textit{latent tokens} for the AdaPerceiver model.
    \textit{N.B.} that AdaPerceiver ($t$=256) nearly matches the mIoU of ViT-H with over 26$\times$ lower FLOPs.
    }
    \label{tab:seg-eval}
    \setlength{\tabcolsep}{6.4pt}
    \begin{tabular}{lr|rrr}
        \toprule
        \textbf{Model} & \textbf{Tokens} & \textbf{Mean IoU} \blueuparrow & \textbf{GFLOPs} \bluedownarrow \\
        \midrule
        ViT-B/32~\cite{reproduciblescalingclip2024} & 256 & 32.5 & 111 \\
        ViT-B/16~\cite{reproduciblescalingclip2024} & 1024 & 39.6 & 437 \\
        ViT-L/14~\cite{reproduciblescalingclip2024} & 1369 & 43.2 & 2071 \\
        ViT-H/14~\cite{reproduciblescalingclip2024} & 1369 & 44.2 & 4313 \\
        \midrule
        \multirow{5}{*}{\textit{AdaPerceiver}} 
        & 32  & 38.4 & 73 \\
        & 64  & 40.5 & 85 \\
        & 96  & 41.8 & 97 \\
        & 128 & 42.3 & 109 \\
        & 192 & 43.3 & 134 \\
        & 256 & 43.9 & 158 \\
        \bottomrule
    \end{tabular}
\end{table}

\begin{figure}[!tb]
    \centering\includegraphics[width=1\linewidth]{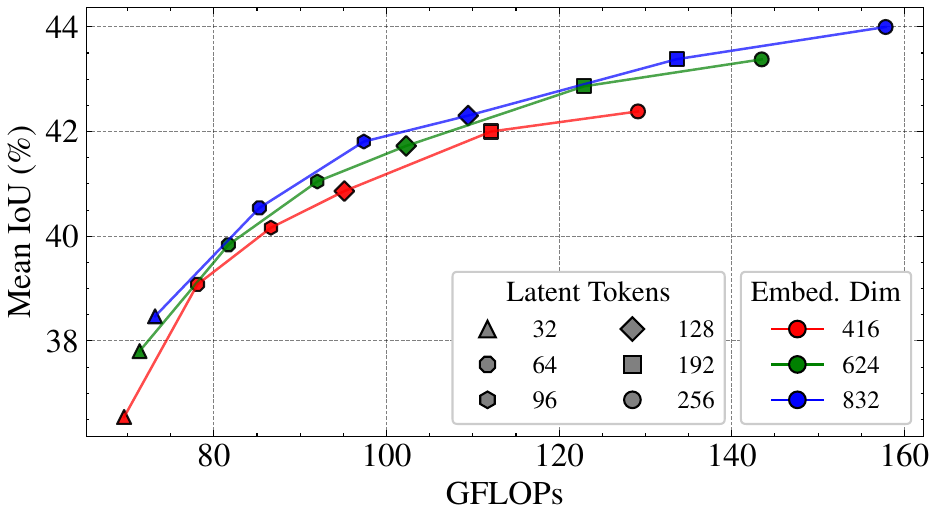}
    \caption{
    \textbf{ADE20K Configuration Tradeoffs.} Mean IoU vs. GFLOPs (encoder) for AdaPerceiver under varying embedding dimensions and latent tokens.
    }
    \label{fig:eval:seg:token-matdim-tradeoff}
\end{figure}

\paragraph{Results}
\cref{tab:seg-eval} summarizes results on ADE20K semantic segmentation.
AdaPerceiver nearly matches its teacher in mIoU while being substantially more efficient. 
With 256 latent tokens, it reaches 43.9 mIoU, 0.3 below ViT-H/14 while using over 26$\times$ fewer FLOPs (158 vs.\ 4313 GFLOPs).
Compared with models with similar computational costs, AdaPerceiver consistently outperforms.

As shown in \cref{fig:eval:seg:token-matdim-tradeoff}, increasing latent tokens or embedding dimension improves performance smoothly, illustrating controllable trade-offs between accuracy and efficiency.
See~\cref{fig:appendix:seg:token-depth} for token-depth trade-offs.

\subsubsection{Depth Estimation}
\label{sec:eval:dense:depth}
We evaluate on the NYUv2 Depth Estimation dataset~\cite{nyuv2_depth}.

\paragraph{Training}
As with semantic segmentation, we use the \textit{linear} head setup from \cite{tips_paper, oquab_dinov2_2023}.
Then, for both AdaPerceiver and the baseline, we upsample patch features by a factor of 4, attach a linear layer to network, upsample the logit predictions to the input resolution, and following \cite{bhat2021adabins} predict depths over 256 uniformly distributed bins.
For AdaPerceiver, we keep the MLP adapter used during feature distillation and attach the linear head.

\begin{table}[!tb]
\centering
\caption{
\textbf{Depth Estimation Evaluation.}
RMSE and Forward FLOPs (encoder only).
For AdaPerceiver, the number of \textit{output tokens is 1369} (matching ViT-L and -H).
We vary the number of  \textit{latent tokens} for AdaPerceiver.
}
\setlength{\tabcolsep}{8.2pt}
\begin{tabular}{lr|rrr}
\toprule
\textbf{Model} & \textbf{Tokens} & \textbf{RMSE} \bluedownarrow & \textbf{GFLOPs} \bluedownarrow \\
\midrule
ViT-B/32~\cite{reproduciblescalingclip2024} & 265 & 0.705 & 115 \\
ViT-B/16~\cite{reproduciblescalingclip2024} & 1064 & 0.693 & 454 \\
ViT-L/14~\cite{reproduciblescalingclip2024} & 1369 & 0.605 & 2082 \\
ViT-H/14~\cite{reproduciblescalingclip2024} & 1369  & 0.585 & 4335 \\
\midrule
\multirow{5}{*}{\textit{AdaPerceiver}} 
& 32  & 0.617 & 73 \\
& 64  & 0.606 & 85 \\
& 96  & 0.592 & 97 \\
& 128 & 0.587 & 109 \\
& 192 & 0.582 & 134 \\
& 256 & 0.579 & 158 \\
\bottomrule
\end{tabular}
\label{tab:depth-eval}
\end{table}
\begin{figure}[!t]
    \centering\includegraphics[width=0.99\linewidth]{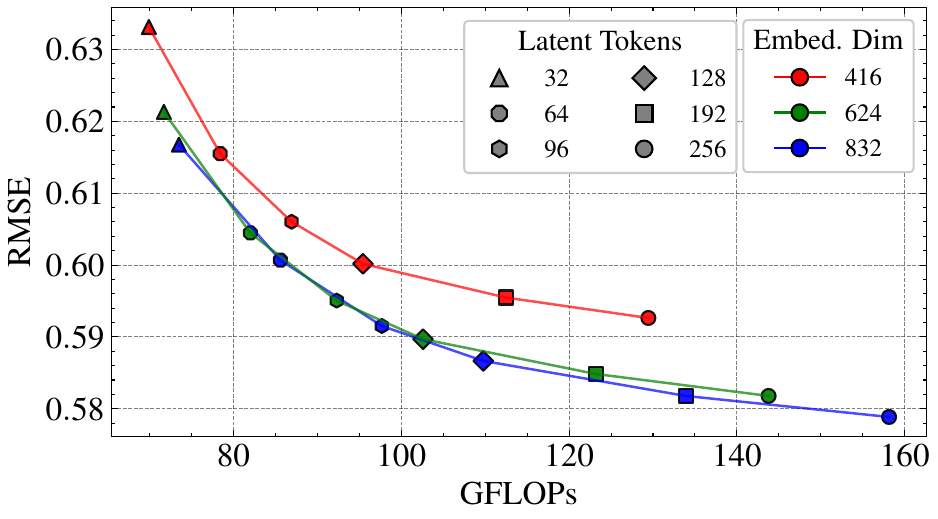}
    \caption{
    \textbf{Depth Estimation Configuration Tradeoffs.}
    RMSE vs. GFLOPs (encoder) for AdaPerceiver under varying embedding dimensions and latent tokens.
    }
    \label{fig:eval:depth:token-matdim-tradeoff}
\end{figure}

\paragraph{Results}
\cref{tab:depth-eval} summarizes results on depth estimation.
At 256 tokens, AdaPerceiver achieves near-equal RMSE to ViT-H/14 while using 96\% fewer FLOPs. 
Furthermore, at 192 tokens it has lower RMSE than all other ViT variants, while using only 134 GFLOPs, which is only 14\% higher than the minimal ViT-B/32 and 96\% lower than ViT-H/14.

\cref{fig:eval:depth:token-matdim-tradeoff} depicts the token-width trade-offs for depth estimation.
Unlike segmentation, width plays are a substantial role in depth estimation.
Substantial improvements in RMSE come from increasing width from 416 to 624.
Further increasing width does not yield comparable gains --- \textit{cf.} \cref{fig:appendix:depth-est:token-depth} for token-depth trade-offs.

\subsection{Qualitative Evaluations}
\label{sec:eval:qual}
\begin{figure}
    \centering
    \includegraphics[width=0.80\linewidth]{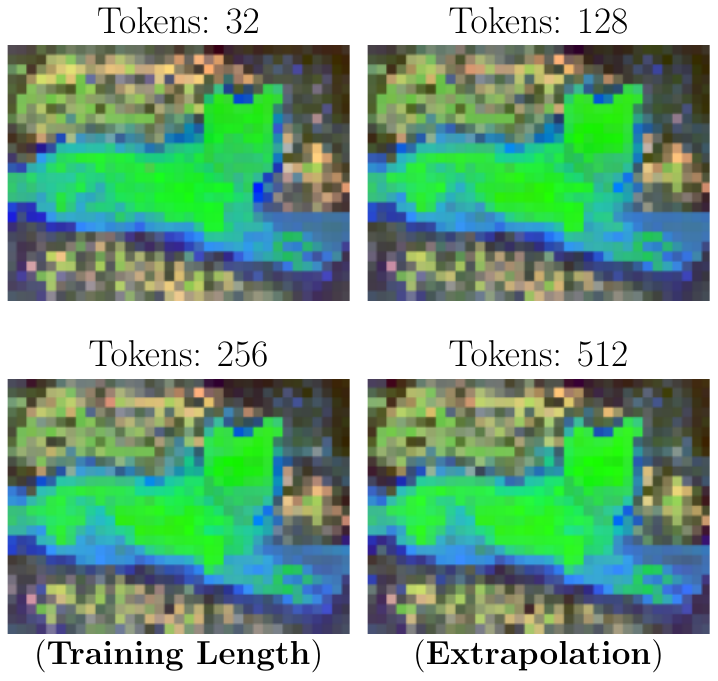}
    \caption{
    First three principal components of the patch features from AdaPerceiver when varying the number of latent tokens (the embedding dimension and depth fixed to their respective maximums). 
    The top three principal components remain consistent across token counts (32 $\rightarrow$ 512).
    The principal components \textbf{remain stable when extrapolating past the training length}.
    }
    \label{fig:eval:qual:tokens}
\end{figure}

\begin{figure*}[t]
    \centering
    \begin{subfigure}[b]{0.45\linewidth}
        \centering
        \includegraphics[width=\linewidth]{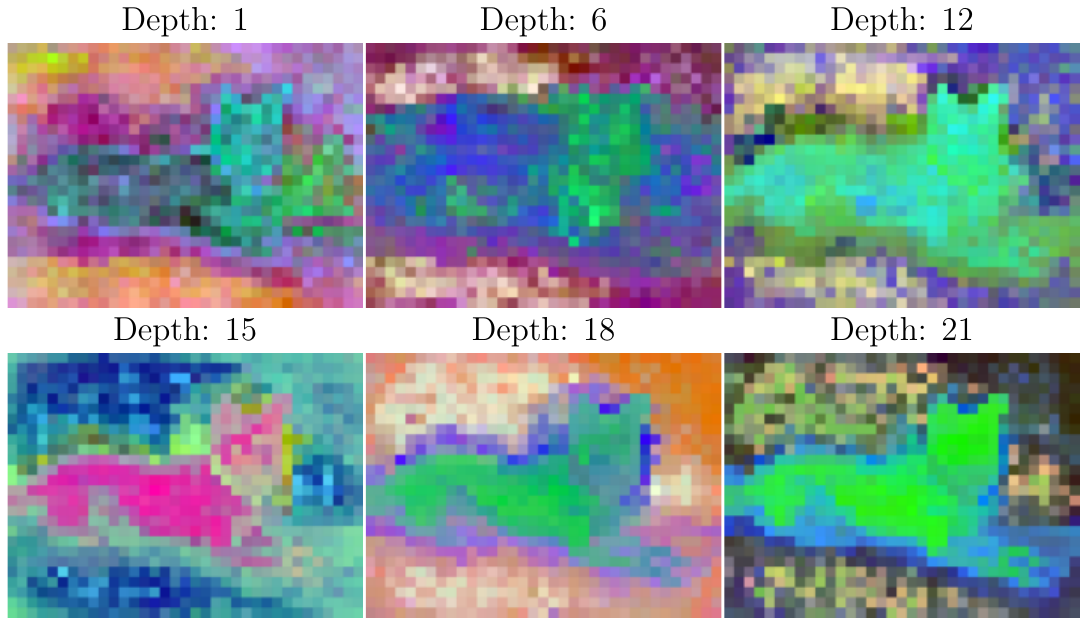}
        \label{fig:appendix:feat-viz:depth_vary:1}
    \end{subfigure}
    \hfill
    \hspace{1.5em}
    \begin{subfigure}[b]{0.45\linewidth}
        \centering
        \includegraphics[width=\linewidth]{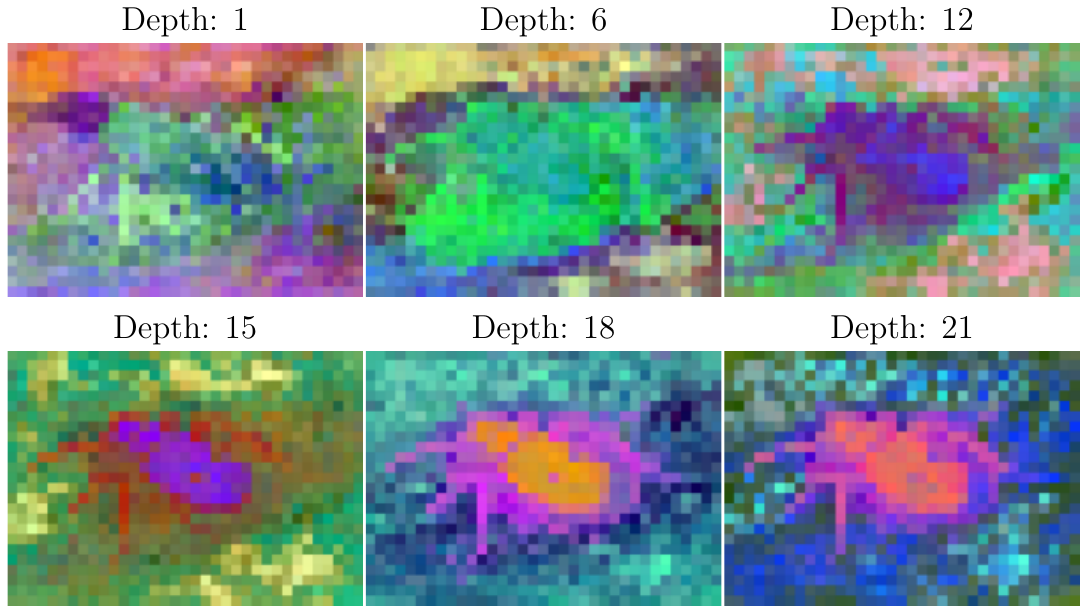}
        \label{fig:appendix:feat-viz:depth_vary:2}
    \end{subfigure}
    
    \caption{
    First three principal components of patch features across depth in AdaPerceiver.
    Discernible semantic features emerge at greater depths, differing per-image.
    Appendix~\cref{fig:appendix:feat-viz:depth_vary} shows patch features at all depths.
    }
    \label{fig:eval:qual:depth}
\end{figure*}

To understand how AdaPerceiver's features change across configurations we depict patch features (principal components) across token granularities (\cref{fig:eval:qual:tokens}) and depth (\cref{fig:eval:qual:depth}).

\cref{fig:eval:qual:tokens} shows that patch features are consistent across the range of supported token granularities ($32\to256$).
Moreover, when extrapolating beyond the trained granularities to 512 tokens the features remain consistent.
We further study the extrapolation and interpolation characteristics in Appendix~\cref{fig:appendix:image-class:inter-extrap,fig:appendix:seg:inter-extrap,fig:appendix:depth-est:inter-extrap}.
The observed extrapolation behaviour is consistent with our expectations from the use of RoPE~\cite{su2024roformer} (\textit{cf.} \cref{sec:meth:arch:latent}).

\cref{fig:eval:qual:depth} illustrates that patch features change significantly across depth. 
The components of particular images stabilize at varying depths: the dog (left) becomes coherent around depth 12, whereas the beetle image of \cref{fig:eval:qual:depth} (right) becomes coherent around depth 15.

\subsection{Policies for Adaptivity}
\label{sec:eval:policies}
AdaPerceiver exposes a large space of valid configurations across tokens, depth, and width.
A configuration must be chosen. 
We therefore evaluate a set of \textit{policies} that govern the choice of configuration and analyze their impact on accuracy-efficiency trade-offs, as shown in~\cref{tab:results:policy}.

Concretely, we study configuration policies on the ImageNet-1K classification task, with the experimental setup outlined in \cref{sec:eval:experimental}. 
We restrict attention to simple policies.
These are detailed in~\cref{sec:appendix:policies}, but briefly:
\begin{itemize}
    \item \textbf{Baseline Policy}: This policy uses a single configuration regardless of input.
    We report policies that choose $t$, the number of tokens, \textit{a priori}.
    \item \textbf{Early Exit (EE) Policy}: This policy combines the Baseline Policy with an early-exit confidence threshold $\tau$~\cite{jiang2024tracing}.
    \item \textbf{RL Policy}: We train a lightweight policy network using REINFORCE~\cite{williams1992simple} to select a token count for an input.
    \item \textbf{Optimal Policy}: To characterize the theoretical upper bound on performance, we define an oracle ``optimal'' policy.  
    Given a trained model, this policy chooses, for each input, the configuration with the least compute that still yields a correct classification.
\end{itemize}

\paragraph{Results}
\cref{tab:results:policy} summarize the effect of different policies for choosing AdaPerceiver configurations (full results in~\cref{sec:appendix:policies}).
The \textit{baseline} policy follows the trade-off shown in \cref{fig:eval:image-class:token-matdim-tradeoff}.
Combining a fixed token count with early exiting can yield ``free lunches'': for example, with 256 tokens and an exit threshold of 0.95 ($\tau$), there is no accuracy degradation with a $\sim$24\% reduction in FLOPs.
The FLOPs reduction increases to $\sim$33\% with only minor degradation at $\tau$=0.9 ($-0.1$ percentage points).
Our RL Policy provides further improvements over early exiting, achieving a reduction of $\sim$8\% FLOPs compared to the 192-token configuration with a 0.9 exit threshold, at a comparable 
accuracy.

\begin{table}[ht!]
\centering
\caption{
\textbf{Adaptivity Policy Evaluation.}
Accuracy and computational cost (GFLOPs) for configuration selection policies applied to AdaPerceiver on image classification.
Combining early-exiting with token reduction proffer ``free-lunches".
\textit{N.B.} The ``Optimal" policy is impractical to realize.
}
\setlength{\tabcolsep}{5.2pt}
\begin{tabular}{lrr}
\toprule
\textbf{Policy} & \textbf{Accuracy} (\%) \blueuparrow & \textbf{GFLOPs} \bluedownarrow \\
\midrule
Baseline ($t=96$) & 84.5 & 40.4 \\
Baseline ($t=128$) & 85.0 & 52.5 \\
Baseline ($t=192$) & 85.3 & 76.7 \\
Baseline ($t=256$) & 85.4 & 100.8 \\
\midrule
EE ($t=128, \tau=0.90$) & 84.7 & 35.0 \\
EE ($t=192, \tau=0.90$) & 85.1 & 51.2 \\
EE ($t=256, \tau=0.90$) & 85.3 & 66.8 \\
EE ($t=256, \tau=0.95$) & 85.4 &  76.5 \\
\midrule
RL (tokens, $\tau=0.90$) & 85.0 & 46.9  \\
\midrule
Optimal & 93.6 & 32.5 \\
\bottomrule
\end{tabular}
\label{tab:results:policy}
\end{table}

\section{Limitations}
\label{sec:limitations}
First, training adaptive models is challenging.
We rely on distillation in this paper to ease learning and to mitigate training costs and do not explore training the model entirely from scratch.
This limits the generality of our approach when high-quality pre-trained teachers are unavailable.

Second, our efficient joint training regime has high memory costs --- though superior to naive joint optimization.

Third, for dense-prediction tasks, we evaluated with a linear probe rather than a state-of-the-art decoder.
As a result, AdaPerceiver's upper-bound performance on these tasks remains unclear.

\section{Conclusion}
\label{sec:conclusion}
We introduce AdaPerceiver, an adaptive architecture that is runtime-configurable along depth, tokens, and width axes.
Specifically, we introduce a novel variant of the Perceiver architecture, and a once-for-all training regime that enables joint-training across these axes.
Our results illustrate that AdaPerceiver outcompetes other adaptive architectures and baselines across classification, semantic segmentation, and depth estimation.
Moreover, because our architecture is configurable at inference time, users can select configurations for their use-cases based on their accuracy/latency requirements.
Efficient learning in adaptive models remains a promising direction for future work.

{
    \small
    \bibliographystyle{ieeenat_fullname}
    \bibliography{bib/main, bib/references}
}

\appendix
\clearpage
\setcounter{page}{1}
\maketitlesupplementary

\section*{Table of Contents}
\label{sec:appendix:toc}
\begin{itemize}
    \item \cref{sec:appendix:rw}: Extended Related Work. 
    \item \cref{sec:appendix:flexi}: Commentary on FlexiViT.
    \item \cref{sec:appendix:architecture}: Architectural Details.
    \item \cref{sec:appendix:training}: Training Details.
    \item \cref{sec:appendix:pseudo}: Pseudocode.
    \item \cref{sec:appendix:image-class}: Image Classification Results.
    \item \cref{sec:appendix:dense}: Dense Prediction Results.
    \item \cref{sec:appendix:feat-viz}: Feature Visualizations.
    \item \cref{sec:appendix:policies}: Policies for Adaptivity.
\end{itemize}

\section{Extended Related Work}
\label{sec:appendix:rw}
We extend the related work presented in \cref{sec:rw}.
In particular, we add some coverage of recent works in \textit{recursive reasoning} models (or alternatively compute scaling models).

\subsection{Adaptive Models}
Adaptivity in deep learning has been explored through two main traditions: dynamic (\ie conditional) neural networks (NNs)~\cite{bengio2015conditional, han2021dynamic}, and elastic models~\cite{cai2019once, devvrit2024matformer} --- however there are recent work in recursive reasoning models have emerged~\cite{trm_2025, scaling_latent_looped_2025, parallel_loop_xformer_2025, scaling_latent_looped_2025}.
In our view, all three can be seen in a unified light and represent multiple paths prior work attempts to reach a common (often implicitly stated) goal.

\paragraph{Dynamic Neural Networks}
Dynamic neural networks adapt computation on a per-input basis, allocating compute or parameters depending on the difficulty or content of the input.
Approaches include early-exiting strategies~\cite{teerapittayanon2016branchynet, wang2018skipnet, wolczyk2021zero, zhou2020bert, raposo2024mixture} and pruning techniques~\cite{gao2018dynamic, xu2022evo, rao2021dynamicvit, bolya2022token, yin2022vit}.
These methods generally either use a heuristic or learn a notion of ``importance'' to decide whether to execute an additional layer (adaptive depth or early-exit), drop tokens (token pruning), or mask features.

\paragraph{Elastic Models}
In contrast, the elastic model tradition focuses on training a \textit{single model} that can be executed at multiple capacities under user-defined compute budgets~\cite{devvrit2024matformer, valipour2023sortednet, haberer2024hydravit, cai2024flextron, hou2020dynabert}. 
Early work such as Once-for-All networks~\cite{cai2019once} demonstrated that convolutional networks can be trained to support a set of sub-networks that trade accuracy for efficiency at inference time. 
Subsequent work extends this idea to Transformer architectures, enabling flexible inference across 1--2 dimensions: tokens, depth, or width. 
Width-adaptive models such as MatFormer, HydraViT, and Flextron~\cite{devvrit2024matformer, valipour2023sortednet, haberer2024hydravit, cai2024flextron} train shared-weight sub-networks that operate at varying hidden dimensions, while DynaBERT and SortedNet~\cite{hou2020dynabert, valipour2023sortednet} explore joint width–depth adaptivity. 
Token-adaptivity has been studied in FlexiViT~\cite{beyer2023flexivit}, which supports varying patch sizes at inference---and thus token counts---within a single model.

Existing training strategies for these models are either costly (relying on multiple forward-passes per configuration~\cite{devvrit2024matformer}) or noisy (stochastic training approaches that sample configurations~\cite{haberer2024hydravit, valipour2023sortednet, cai2024flextron, beyer2023flexivit}).

\paragraph{Recursive Reasoning Models}
Compared with elastic models and (some) dynamic neural networks, recursive reasoning models~\cite{parallel_loop_xformer_2025,scaling_latent_looped_2025,geiping2025scaling, trm_2025} (and the older Universal Transformers~\cite{dehghani2018universal}) recur over a core architectural block to solve predictive tasks.
This is meant as a means of scaling ``test-time compute"~\cite{geiping_scale_test_time_2025}, and often is coupled with some halting condition~\cite{scaling_latent_looped_2025}.

We refrained from introducing this work in the main body because although many of these models could be considered adaptive, they are not necessarily comparable to dynamic neural networks and elastic models. 
Elastic models \textit{budget} the compute of a model, whereas these reasoning models \textit{expand} the base compute.
In abstract, if a model expends a single compute unit, then elastic methods attempt to partition that unit to expose models capable of expending various amounts of compute (up to a maximum).
Whereas recursive reasoning models can repeatedly expend this unit of compute.

\paragraph{Comparison to Our Work}
AdaPerceiver combines elements of both dynamic neural networks and elastic model.  
Like dynamic neural networks, it supports per-input adaptivity: configurations can be selected at runtime, \eg by a learned policy (see \cref{sec:eval:policies}).  
Similar to elastic models and reasoning models, we train a single shared-weight model to support flexible configurations.  
However, unlike prior elastic models, AdaPerceiver supports simultaneous adaptivity across token, depth, and width axes.  
For our novel training approach, %
we structure the network such that multiple configurations can be \textit{jointly optimized within a single forward pass}.
Training AdaPerceiver does not require multiple forward evaluations, with less reliance on stochastic configuration sampling.

\subsection{Perceiver Architectures}
\label{sec:appendix:rw:perceiver}
Perceiver architectures follow an \textit{encode-process-decode} paradigm:
  inputs are \textit{encoded} via attention into a fixed set of latent tokens (the latent stream);
  this latent stream is \textit{processed} through iterative transformer layers;
  and finally
  \textit{decoded} to produce outputs. 
The original Perceiver introduced a fixed-size latent stream that decoupled input size from internal computation, enabling scalability to large and multi-modal data~\cite{perceiverjaegle21a}.
PerceiverIO extended this idea by introducing an output query mechanism, allowing latent representations to be decoded into arbitrarily sized outputs~\cite{jaegle2021perceiverio}. 
Subsequent variants further developed this direction. PerceiverAR~\cite{hawthorne2022percevierar} adapted the architecture for autoregressive modeling, while the Hierarchical Perceiver (HiP)~\cite{carreira2022hip} incorporated locality and hierarchical structure to improve efficiency while maintaining generality.

\paragraph{Comparison to Our Work}
Prior work on the Perceiver family studies generality and scalability across modalities.
Their latent processing streams are fixed once trained. 
We introduce adaptivity into this latent stream, enabling control over the amount of computation allocated to each input.

\section{Why \textit{not} FlexiViT for token adaptivity?}
\label{sec:appendix:flexi}
FlexiViT is a plausible means of achieving token adaptivity, and one may naturally ask why a Perceiver-style architecture is required. 
In principle, one could combine early exiting, Matryoshka learning, and FlexiViT within a standard ViT and obtain adaptivity along all three axes.
We provide a summary of FlexiViT in \cref{sec:appendix:flexi:summary}, and our reasons for not using it \cref{sec:appendix:flexi:reasons}.

\subsection{Summary of FlexiViT}
\label{sec:appendix:flexi:summary}
FlexiViT achieves token adaptivity by varying the patch size used to encode the input.
Smaller patch sizes yield more tokens (and thus more compute), whereas larger patch sizes yield fewer tokens (and lower compute).
During training, FlexiViT samples patch sizes uniformly on a per-batch basis, allowing the patch size---and by extension the compute budget---to be adjusted at inference.

Compared to AdaPerceiver, FlexiViT achieves token adaptivity by varying the patch size, whereas AdaPerceiver does so by changing the number of tokens in the latent stream (\textit{cf.}~\cref{sec:appendix:rw:perceiver,sec:meth:arch,sec:appendix:architecture}). 

\subsection{Reasons not to use FlexiViT}
\label{sec:appendix:flexi:reasons}

In this work, we do not use FlexiViT for two reasons.

\begin{figure}
    \centering
    \includegraphics[width=1.0\linewidth]{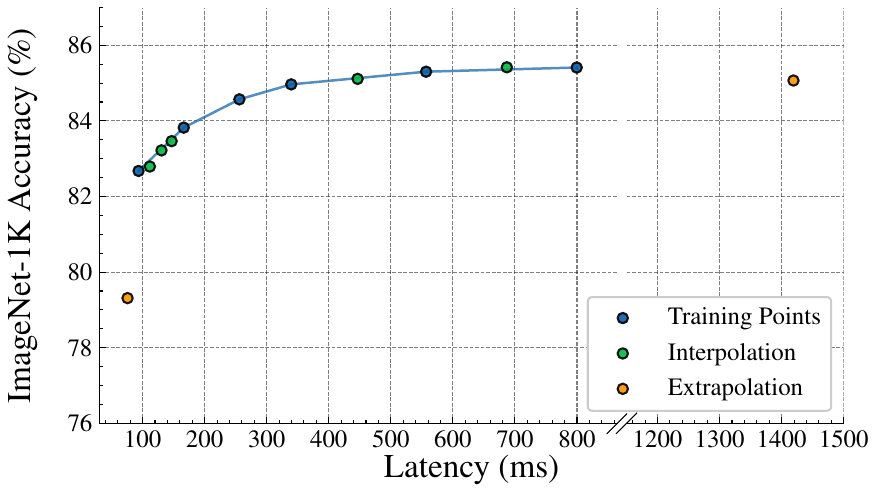}
    \caption{\textbf{
    Effect of Latent Token Interpolation and Extrapolation on ImageNet-1K Accuracy.} 
    AdaPerceiver is trained with token granularities, $\mathcal{T} = \{32, 64, 96, 128, 192, 256\}$, however it able to interpolate (\textcolor{ForestGreen}{green} points) within $\mathcal{T}$ and extrapolate outside $\mathcal{T}$ (\textcolor{Dandelion!90!red}{yellow} points).
    When interpolating, AdaPerceiver remains on the Pareto frontier, whereas extrapolation leads to some degradation in accuracy, with the largest drop occurring when extrapolating below the smallest trained token granularity.
    \textit{N.B. The x-axis contains a break to ease visualization.}
    }
    \label{fig:appendix:image-class:inter-extrap}
\end{figure}

\begin{figure}
    \centering
    \includegraphics[width=1.0\linewidth]{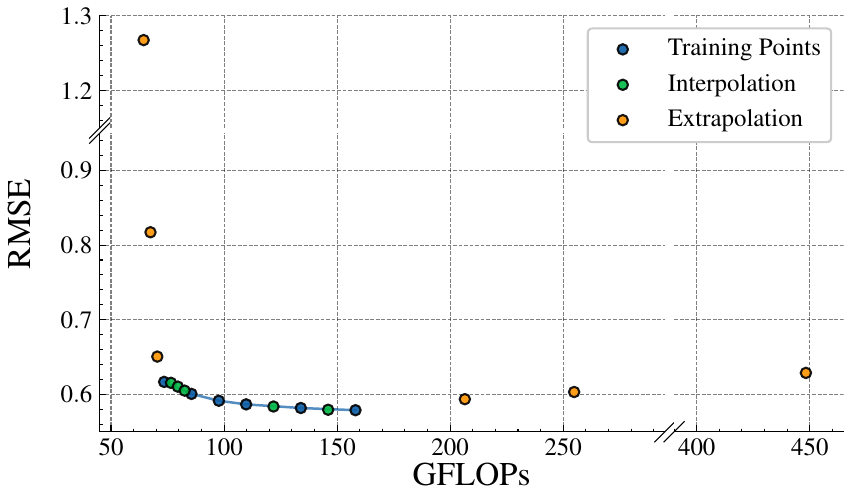}
    \caption{
    \textbf{Effect of Latent Token Interpolation and Extrapolation on NYUv2 depth estimation.} 
    Similar to \cref{fig:appendix:image-class:inter-extrap}, AdaPerceiver is able to interpolate however it able to interpolate (\textcolor{ForestGreen}{green} points) between its training token granularities and to extrapolate (\textcolor{Dandelion!90!red}{yellow} points) beyond them.
    Performance degradation appears when extrapolating outside the trained range, with the largest drop occurring when extrapolating below the smallest trained token granularity.
    \textit{N.B. Both the x-axis and y-axis contain breaks to ease visualization.}
    }
    \label{fig:appendix:depth-est:inter-extrap}
\end{figure}

\paragraph{Limited Control Over Token Count}
FlexiViT does not support arbitrary token counts.
The number of tokens is determined jointly by the input resolution and the patch size.
Thus, changing the patch size does \textit{not} guarantee the same number of tokens --- nor the same amount of compute --- across different input sizes.
In contrast, AdaPerceiver can map any input resolution to an arbitrary number of latent tokens.
Moreover, it supports interpolation and extrapolation beyond the token granularities used during training (\cref{fig:appendix:image-class:inter-extrap,fig:appendix:depth-est:inter-extrap,fig:appendix:seg:inter-extrap}).
We note that this flexibility is a design trade-off: in some settings, scaling compute with input resolution is desirable.
AdaPerceiver does not automatically scale compute with increasing input resolution, as the number of latents is controlled independently from input size.

\paragraph{Input and Output Token Counts Are Coupled}
A more subtle limitation concerns dense prediction.
In FlexiViT, the number of input tokens equal to the number of output tokens: reducing input tokens necessarily reduces output tokens.
However, token count is known to strongly influence dense prediction performance~\cite{wang2025scaling}.
Ideally, one would like to process \emph{fewer} tokens (for efficiency) while still producing \emph{more} output tokens (for predictive performance).

AdaPerceiver supports this decoupling: it can process a lower number of latent tokens while decoding to a higher number of output tokens.
Our results suggest that processing fewer tokens while outputting more can yield performance comparable to processing more tokens (\textit{cf.} \cref{tab:seg-eval,tab:depth-eval}), though further investigation is needed.

If this effect is real, then FlexiViT fundamentally lacks the ability to exploit it, as its output token count is inherently tied to its input token count.

\section{Architectural Details}
\label{sec:appendix:architecture}
We elaborate upon our architecture details in \cref{sec:meth:arch}.
\subsection{Notation}
\label{sec:appendix:architecture:notation}
We denote by 
$x \in \mathbb{R}^{I \times d_i}$ the sequence of $I$ input tokens with embedding dimension $d_i$,
by $z_l \in \mathbb{R}^{N \times d_z}$ the $N$ latent tokens at layer $l \in [0, L]$ with embedding dimension $d_z$,
and by $o \in \mathbb{R}^{M \times d_o}$ the $M$ output tokens with embedding dimension $d_o$.  
The learned latent token is denoted $z \in \mathbb{R}^{1 \times d_z}$, and $z_0$ is the initial latent array obtained after broadcasting and reading from the input.

We define three sets: 
$\mathcal{T}$ for token granularities, 
$\mathcal{W}$ for width configurations, and 
$\mathcal{D}$ for depths. 
Each adaptive sub-network is indexed by a tuple $(t, w, l)$ where 
$t \in \mathcal{T}$, $w \in \mathcal{W}$, and $l \in \mathcal{D}$.

\subsection{AdaPerceiver Architecture}
\label{sec:appendix:architecture:arch-details}
AdaPerceiver follows the \textit{encode–process–decode} paradigm of Perceiver (\textit{cf.}~\cref{fig:method:architecture}).

\paragraph{Encode}
We first broadcast the learned latent token $z$ to $N$ latents, producing $z' \in \mathbb{R}^{N \times d_z}$:
\begin{equation}
    z' = \mathtt{Broadcast}(z,\, N) = [z, z, \ldots, z] \in \mathbb{R}^{N \times d_z}.
\end{equation}

We then read the input tokens into this broadcast latents using cross-attention, treating $z'$ as sink (query) tokens and the input tokens $x$ as source (key/value) tokens:
\begin{equation}
    z_0 = z^\prime + \mathtt{CrossAttention}(z',\, x).
\end{equation}
We apply RoPE~\cite{su2024roformer} to the sink tokens $z'$ to positionally distinguish the tokens.

The \textit{encode} step assumes we are given input tokens $x$.
In practice, the input tokens are modality specific.
For images, they may be patches or features of a network.
We elaborate further in \cref{sec:appendix:architecture:input-tokens}.

\paragraph{Process}
Following the encode step, the latents $z_0$ are refined using a sequence of AdaPerceiver blocks:
\begin{equation}
    z_l = \mathtt{AdaPerceiverBlock}(z_{l-1}), \qquad l \in [1, L].
\end{equation}
Each $\mathtt{AdaPerceiverBlock}$ has the same structure as ViT blocks:
\begin{align}
    z_l^{\prime} &= z_{l-1} + \mathtt{BlockMaskAttention}(\mathtt{Norm}(z_{l-1})), \\
    z_l &= z_l^{\prime} + \mathtt{MatFFN}(\mathtt{Norm}(z_l^{\prime})).
\end{align}

We use $\mathtt{Norm}$ to denote LayerNorm.
$\mathtt{BlockMaskAttention}$ refers to multi-head attention with RoPE applied to the queries and keys and block attention mask.
$\mathtt{MatFFN}$ denotes the feed-forward network with Matryoshka linear layers ($\mathtt{MatLinear}$); a minimal $\mathtt{MatLinear}$ implementation is shown in \cref{alg:appendix:matlinear}.

\paragraph{Decode}
Finally, the latent tokens are decoded (read out) to output tokens. 
This step is nearly identical to the \textit{Encode} step.
Given a set of output tokens, $o$, the latent tokens are read to the output tokens using cross-attention, treating the $o$ as the sink (query) tokens and the latent tokens $z_l$ as the source (key/value):
\begin{equation}
    o = o' + \mathtt{CrossAttention}(o',\, z_l).
\end{equation}
Where  $o'$ are the initial output tokens. 
We elaborate on how the output tokens can be initialized for classification and dense prediction tasks in \cref{sec:appendix:architecture:output-tokens}

\subsection{Designing the Latent Token(s)}
\label{sec:appendix:latent}
As noted in \cref{sec:meth:arch:latent}, various choices are available for the latent tokens $z$. We outline two options below.

\paragraph{Learned Latent Array}
PerceiverIO learns $N$ latent tokens equal to the size of their latent stream~\cite{perceiverjaegle21a, jaegle2021perceiverio}.
This works well for fixed latent arrays (as in PerceiverIO) but does not translate well when we may want the latent stream to be adaptive --- in such cases one could up-sample the latent array when requiring $>N$ tokens or down-sample when wanting $<N$.
However, this is simply more complicated than learning a single token broadcast to $N$, then applying RoPE (as we do in \cref{sec:meth:arch:latent}).
    
\paragraph{Randomly Initialized Latents}
A recent work by Geiping \etal follows the \textit{encode-process-decode} scheme of Perceiver models, in their case they initialize their latent tokens from a normal distribution $\mathcal{N}(0, \sigma I)$~\cite{geiping2025scaling}.
Such a scheme can be used in lieu of learning a single token by just sampling a token vector from $\mathcal{N}(0, \sigma I)$, or an array of latents can be sampled.
We mention this for sake of completeness but did not study it further.

\subsection{Why block masking allows for adaptive tokens?}
\label{sec:appendix:block-mask}

To understand why block masking enables adaptive token granularities, it is useful to examine the attention layer within the $\mathtt{AdaPerceiverBlock}$, as this is the only component that performs sequence-level mixing (\textit{c.f.,}~\cite{hwang2024hydra}).  
In standard attention, the layer forms a mixing matrix $A \in \mathbb{R}^{N \times N}$ from the queries and keys $Q, K \in \mathbb{R}^{N \times d}$, and applies it to the values $V \in \mathbb{R}^{N \times d}$:
\begin{equation*}
    Y = A V,
\end{equation*}
so that each output token $y_i$ is a weighted combination of all value tokens:
\begin{equation*}
    y_i = \sum_{j=1}^{N} a_{ij} v_j.
\end{equation*}

With block masking, we constrain which tokens may interact during attention.  
Specifically, a token may only mix with tokens from its own granularity and with those from \textit{smaller} granularities.  
For example, consider training an AdaPerceiver model with two token granularities $\mathcal{T} = \{2, 4\}$.  
The block mask enforces:
\begin{align*}
    y_i &= 
    \begin{cases}
        \sum_{j=1}^{2} a_{ij} v_j, & i \in \{1, 2\}, \\
        \sum_{j=1}^{4} a_{ij} v_j, & i \in \{3, 4\}.
    \end{cases}
\end{align*}

Thus, the first two output tokens depend exclusively on the first two input tokens, while the last two depend on all four.  
Because the first two outputs only result from a mixing of the first two inputs, their computation is \textit{identical} to the computation the model would perform if the sequence length were actually two.  
In general, block masking ensures that the computation associated with any granularity depends only on the tokens belonging to that granularity and the granularities smaller than it.  
Consequently, adding additional tokens (granularities) does not alter the computation of earlier ones, and supervising each granularity during training is therefore equivalent to training the model with multiple numbers of latent tokens.

\subsection{Input Tokens}
\label{sec:appendix:architecture:input-tokens}
We obtain input tokens using the standard patch embedding used in Vision Transformers~\cite{dosovitskiy2020image}.  
Other choices are possible---for example, a smaller pre-trained model or convolutional stem can also be used to produce the input token sequence.

\subsection{Output Tokens}
\label{sec:appendix:architecture:output-tokens}
We describe how output tokens are instantiated for both classification and dense prediction tasks.

\paragraph{Classification}
For classification, we simply learn a single output token.

\paragraph{Dense Prediction}
For dense prediction tasks, we consider two cases.  
If the number of output tokens is known \textit{a priori}, we can directly learn that number of output tokens.  
However, when the number of output tokens is unknown or should scale with the input resolution, we initialize the output tokens from the input tokens themselves.  
In our work, we adopt this latter approach: the output tokens are obtained by applying a learned linear projection to the input tokens.

This construction is beneficial for dense prediction (and feature distillation) because the number of output tokens automatically grows with the number of input tokens (e.g., when increasing image resolution).  
As a result, the model exhibits shape behaviour similar to a traditional ViT.  
Furthermore, this design enables variable-resolution training without having to re-learn or interpolate a fixed set of output latents (a common trick in ViT training).

\subsection{Model Details}
We summarize the key architectural hyperparameters used in our AdaPerceiver model in the table below.  
\begin{table}[h]
\centering
\small
\begin{tabular}{lr}
\toprule
\multicolumn{2}{c}{\textbf{Model Configuration}} \\
\midrule
$\mathcal{W}$ & $\{416, 624, 832\}$ \\
$\mathcal{T}$ & $\{32, 64, 96, 128, 192, 256\}$\\
$\mathcal{D}$ & $\{1,2,\ldots,21\}$\\
\midrule
Input Adapter & Patch Embed. \\
Image Size & $224$ \\
In Channels & $3$ \\
Patch Size & $14$ \\
Embed FFN & \checkmark \\
\midrule
Embed. Dim & $832$ \\
FFN Ratio & $2.57$ \\
Heads & $13$ \\
Depth & $21$ \\
Max Latent Tokens & $256$ \\
RoPE Theta & $10000$ \\
QKV Bias & \checkmark \\
Proj. Bias & \checkmark \\
Layer Scale Init & $1.0 \times 10^{-5}$ \\
\midrule
FFN Activation & GeLU \\
Norm Layer & LayerNorm \\
FFN Layer & MLP \\
\bottomrule
\end{tabular}
\end{table}

\clearpage
\section{Training Details}
\label{sec:appendix:training}
We outline our pre-training and fine-tuning details below.

\subsection{Pre-training/Distillation}
We summarize our training setting below.
We first train solely using logit distillation, and then have a subsequent feature distillation stage for dense prediction tasks.
Our teacher model is the ViT-H/14 CLIP model fine-tuned on ImageNet-12k as the teacher, trained in \cite{reproduciblescalingclip2024} and publicly available in \cite{rw2019timm}.
In both cases we conduct distillation with the ImageNet-12K dataset~\cite{timm_imagenet12k_wds}.
We use the same augmentation settings for both logit and feature distillation:
\begin{table}[h!]
\centering
\small
\begin{tabular}{l|r}
\toprule
\multicolumn{2}{c}{\textbf{Augmentations}} \\
\midrule
Image Size & $224$ \\
Horiz. flip & \checkmark \\
RandAugment & \checkmark \\
RandAug Ops & $2$ \\
RandAug Magnitude & $15$ \\
Mixup & \checkmark \\
Mixup $\alpha$ & $1.0$ \\
\midrule
\end{tabular}
\end{table}

\paragraph{Logit Distillation}
We base our distillation recipe on \cite{beyer2022knowledge}.
We train in three stages, we first train adaptivity over the \textit{token} dimension, then jointly over \textit{token} and \textit{depth}, and finally over all three dimensions. 
At the beginning of each stage \textit{we initialize the model weights with the EMA weights from the prior stage}.
We use the following optimization hyper-parameters:
\begin{table}[h!]
\centering
\small
\begin{tabular}{l|c|c|c}
\toprule
 & \textbf{Stage 1} & \textbf{Stage 2} & \textbf{Stage 3} \\
\midrule
Effective Batch Size     & \multicolumn{3}{c}{$4096$} \\
Optimizer                & \multicolumn{3}{c}{Shampoo-Soap} \\
\midrule
Learning Rate            & $1\times 10^{-3}$ & $1\times 10^{-3}$ &  $5\times 10^{-4}$ \\
\midrule
Weight Decay             & \multicolumn{3}{c}{$0.003$} \\
Betas                    & \multicolumn{3}{c}{$(0.9,\; 0.999)$} \\
Grad Clip                & \multicolumn{3}{c}{$3$} \\
\midrule
EMA Decay                & $0.999$ & $0.999$ & $0.9998$ \\
Epochs               & $50$ & $65$ & $20$ \\
\midrule
Schedule                 & \multicolumn{2}{c|}{Cosine} & -- \\
\midrule
Warmup Steps             & \multicolumn{3}{c}{$3000$} \\
Warmup LR                & \multicolumn{3}{c}{$1\times 10^{-6}$} \\
\midrule
Min LR                   & \multicolumn{2}{c|}{$1\times 10^{-5}$} & -- \\
\midrule
Precond.\ Frequency      & \multicolumn{3}{c}{$100$} \\
Max Precond.\ Dim        & \multicolumn{3}{c}{$8{,}192$} \\
Start Precond.\ Step     & \multicolumn{3}{c}{$500$} \\
\bottomrule
\end{tabular}
\end{table}

For each stage we use the following settings for our loss functions:
\begin{table}[!h]
\centering
\small
\begin{tabular}{lccc}
\toprule
 & \textbf{Stage 1} & \textbf{Stage 2} & \textbf{Stage 3} \\
\midrule
Token Loss           & \checkmark & \checkmark & \checkmark \\
Depth Loss           & --         & \checkmark & \checkmark \\
Width Loss           & --         & --         & \checkmark \\
Depth Loss Schedule  & --         & linear     & linear     \\
\bottomrule
\end{tabular}
\end{table}

\noindent\textit{N.B.} We weight the loss from earlier depths lower than that from later depths and linearly increase the weights: 
the contribution from depth $1$ has weight $1/21$, while the contribution from depth $21$ has weight $1.0$.

\paragraph{Feature Distillation}
For feature distillation we configure our model as dense prediction task and attach a MLP adapter to the output tokens to predict the features our teacher.
We base our feature distillation recipe on \cite{ranzinger2024radio}.
Specifically, we use both the cosine similarity loss ($\mathcal{L}_{\text{cos}}$) and smooth L1 magnitude loss ($\mathcal{L}_{\text{norm}}$):
\begin{align*}
    \mathcal{L_\text{feat}} &= w_{\text{cos}} \mathcal{L}_{\text{cos}} + w_{\text{norm}}\mathcal{L}_{\text{norm}}
\end{align*}
We initialize our model using the \textit{Stage 3} weights.
We use the following optimization hyper-parameters:
\begin{table}[h!]
\centering
\small
\begin{tabular}{l|c}
\toprule
& \textbf{Feature Distillation} \\
\midrule
Effective Batch Size & 4096 \\ 
Optimizer & Shampoo-Soap \\ 
Learning Rate & $5\times10^{-4}$ \\ 
Weight Decay & $0.003$ \\ 
Betas & $(0.9, 0.999)$\\ 
Grad Clip & $3$ \\ 
EMA Decay & $0.9995$ \\ 
Epochs & 20 \\ 
Schedule & Cosine\\ 
Warmup Steps & $3000$ \\ 
Warmup LR & $1\times10^{-6}$\\ 
Min LR & $1\times10^{-5}$\\ 
Precond.\ Frequency & $30$ \\ 
Max Precond.\ Dim & $8192$ \\ 
Start Precond.\ Step & $500$ \\ 
\bottomrule
\end{tabular}
\end{table}

We use the following parameters for our loss:
\begin{table}[h!]
\centering
\small
\begin{tabular}{lc}
\toprule
 & \textbf{Feature Distillation} \\
\midrule
Token Loss      & \checkmark \\
Depth Loss      & \checkmark \\
Width Loss      & \checkmark \\
Depth Loss Schedule & linear    \\
$w_{\text{cos}}$ & 0.9 \\
$w_{\text{norm}}$ & 0.1 \\
\bottomrule
\end{tabular}
\end{table}

\clearpage
\subsection{ImageNet-1K Classification Fine-Tuning}
For ImageNet-1K fine-tuning, we fine-tune the write head (cross-attention to output tokens), output tokens, and a final linear projection layer to project the output token to predict 1000 classes.
We use the following data augmentations:
\begin{table}[h!]
\centering
\small
\begin{tabular}{l|r}
\toprule
\multicolumn{2}{c}{\textbf{Augmentations}} \\
\midrule
Image Size & $224$ \\
Horiz. flip & \checkmark \\
RandAugment & \checkmark \\
RandAug Ops & $2$ \\
RandAug Magnitude & $20$ \\
Mixup & \checkmark \\
Mixup $\alpha$ & $0.8$ \\
CutMix & \checkmark \\
CutMix $\alpha$ & $0.5$ \\
Random Erasing & \checkmark \\
Erase Prob. & 0.25 \\
\midrule
\end{tabular}
\end{table}

We use the same loss functions as Distillation \textit{Stage 3} and use the following optimization hyper-parameters:
\begin{table}[h!]
\centering
\small
\begin{tabular}{l|c}
\toprule
& \textbf{IN-1K Fine-Tuning} \\
\midrule
Effective Batch Size & 1024 \\ 
Optimizer & Shampoo-Soap \\ 
Learning Rate & $5\times10^{-4}$ \\ 
Weight Decay & $0.003$ \\ 
Betas & $(0.9, 0.999)$\\ 
Grad Clip & $3$ \\ 
EMA Decay & $0.9995$ \\ 
Epochs & 70 \\ 
Schedule & Cosine\\ 
Warmup Steps & $500$ \\ 
Warmup LR & $1\times10^{-6}$\\ 
Min LR & $1\times10^{-5}$\\ 
Precond.\ Frequency & $30$ \\ 
Max Precond.\ Dim & $8192$ \\ 
Start Precond.\ Step & $250$ \\ 
\bottomrule
\end{tabular}
\end{table}

\clearpage
\subsection{ADE20K Semantic Segmentation Fine-Tuning}
We fine-tune the write head (cross-attention to output tokens), output tokens, and a final linear projection layer.
We only use the token loss and disable adaptivity training for width and depth. 
We train using the following optimization hyper-parameters:
\begin{table}[h!]
\centering
\small
\begin{tabular}{l|c}
\toprule
& \textbf{ADE20K Fine-Tuning} \\
\midrule
Effective Batch Size & $16$ \\ 
Total Steps & $50,530$ \\ 
Optimizer & Shampoo-Soap \\ 
Learning Rate & $5\times10^{-4}$ \\ 
Weight Decay & $0.1$ \\ 
Betas & $(0.9, 0.999)$\\ 
Grad Clip & $3$ \\ 
EMA Decay & $0.9995$ \\ 
Schedule & Poly. \\ 
Poly. Power & $1$\\ 
Warmup Steps & $1500$ \\ 
Warmup LR & $1\times10^{-6}$\\ 
Min LR & $1\times10^{-6}$\\ 
Precond.\ Frequency & $30$ \\ 
Max Precond.\ Dim & $8192$ \\ 
Start Precond.\ Step & $500$ \\ 
\bottomrule
\end{tabular}
\end{table}

\subsection{NYUv2 Depth Estimation Fine-Tuning}
We fine-tune the write head (cross-attention to output tokens), output tokens, and a final linear projection layer.
We only use the token loss and disable adaptivity training for width and depth. 
We train using the following optimization hyper-parameters:

\begin{table}[h!]
\centering
\small
\begin{tabular}{l|c}
\toprule
& \textbf{Depth Fine-Tuning} \\
\midrule
Effective Batch Size & $16$ \\ 
Total Steps & $47,584$ \\ 
Optimizer & NAdamW \\ 
Learning Rate & $1\times10^{-4}$ \\ 
Weight Decay & $0.1$ \\ 
Betas & $(0.9, 0.999)$\\ 
Grad Clip & $3$ \\ 
EMA Decay & $0.9995$ \\ 
Schedule & Poly. \\ 
Poly. Power & $1$\\ 
Warmup Steps & $1500$ \\ 
Warmup LR & $1\times10^{-6}$\\ 
Min LR & $1\times10^{-6}$\\ 
\bottomrule
\end{tabular}
\end{table}

\clearpage
\section{Pseduocode}
\label{sec:appendix:pseudo}
Here, we introduce PyTorch-esque code for our implementation of Matyroshka linear layers (\cref{alg:appendix:matlinear}) and the AdaPerceiver training regime (\cref{alg:appendix:adaperceiver}).

\begin{algorithm}[!h]
\caption{Matryoshka Linear Layer with per-sample masking.}
\algcomment{
    \textbf{Notes:} Each sample $i$ uses a different embedding dimension $w_i$. The layer either masks inputs (\texttt{mat\_input=True}) or outputs (\texttt{False}) before or after the linear projection.
    During test-time we do not need to rely on masking and can just slice the weight matrices as per \cite{devvrit2024matformer}.  
}
\label{alg:appendix:matlinear}
\lstset{style=mocov3, numbers=left, numberstyle=\tiny, numbersep=5pt}
\begin{lstlisting}[language=python,escapechar=@]
class MatLinear(nn.Linear):
  def forward(self, x, mat_dim, mat_input=False):
    # x: (B, T, in_dim); mat_dim: (B,) adaptive width per sample
    B, T, in_dim = x.shape
    out_dim = self.weight.shape[0]
    mat_dim = mat_dim.to(torch.long, device=x.device)

    if mat_input:
      # Mask input features before projection
      col_idx = torch.arange(in_dim, device=x.device)
      mask = (col_idx.unsqueeze(0) < mat_dim.unsqueeze(1)).unsqueeze(1)  # (B,1,in)
      x = x * mask.to(x.dtype)
      y = F.linear(x, self.weight, self.bias)  # (B,T,out)
    else:
      # Projection, then mask outputs
      y = F.linear(x, self.weight, self.bias)  # (B,T,out)
      row_idx = torch.arange(out_dim, device=x.device)
      mask = (row_idx.unsqueeze(0) < mat_dim.unsqueeze(1))               # (B,out)
      y = y * mask.unsqueeze(1).to(y.dtype)                              # (B,T,out)
    return y
\end{lstlisting}
\end{algorithm}
\begin{algorithm}[!h]
\caption{AdaPerceiver Training.}
\label{alg:appendix:adaperceiver}
\lstset{style=mocov3}
\begin{lstlisting}[language=python,escapechar=@,label=code:adaperceiver]
width_choices = [...]
latent_token_grans = [...] # the last entry corresponds to the max latents used during training
mask = create_block_mask(latent_token_grans) # creates structured mask 

class AdaPerceiver(...):
  def forward_training(x, mask, widths):
    latents = ... # [B, N], N corresponds to the maximum latents tokens used during training.
    output_tokens = ... # [B, M], # M corresponds to the output tokens

    # cross attention from input to latents.
    # latents are the sink (Q), x is the source (K, V).
    latents = cross_attention(latents, x)

    # apply adaperceiver blocks to latents
    # we provide the mask and per sample width
    final_latents, intermediate_latents = forward_blocks(latents, mask, widths)

    output_list = []
    intermediate_output_list = []
    for token_gran in latent_token_grans:
      # We select the first token_gran tokens and read them out.
      sliced_latents = latents[:, :token_gran]
      # cross attention from input to latents.
      # output_tokens are the sink (Q), sliced_latents are the source (K, V).
      token_gran_output = cross_attention(output_tokens, sliced_latents)
      output_list.append(token_gran_output)

    for int_latent in intermediate_latents:
      # we sample a token granularity and slice the latents.
      token_gran = sample(latent_token_grans)
      sliced_latents = latents[:, :token_gran]
      # cross attention from input to latents.
      # output_tokens are the sink (Q), sliced_latents are the source (K, V).
      token_gran_output = cross_attention(output_tokens, sliced_latents)
      intermediate_output_list.append(token_gran_output)
    return output_list, intermediate_output_list

model = AdaPerceiver(...)
for x, y in dataloader:
  B = x.shape[0] # get the batch size
  # Sample width for each sample in the batch
  widths  = [sample(width_choices) for _ in range(B)] 

  # Forward and backward pass
  output_list, int_output_list = model.forward_training(x, mask, widths)

  # token loss
  token_loss = loss_fn(output_list, y)
  # layer_loss
  layer_loss = loss_fn(int_output_list, y)
  loss = token_loss + layer_loss
  loss.backward()
  ...
\end{lstlisting}
\end{algorithm}

\clearpage
\section{Image Classification Results}
\label{sec:appendix:image-class}
\begin{figure*}[ht]
    \centering
    \begin{subfigure}[t]{0.48\linewidth}
        \centering
        \includegraphics[width=\linewidth]{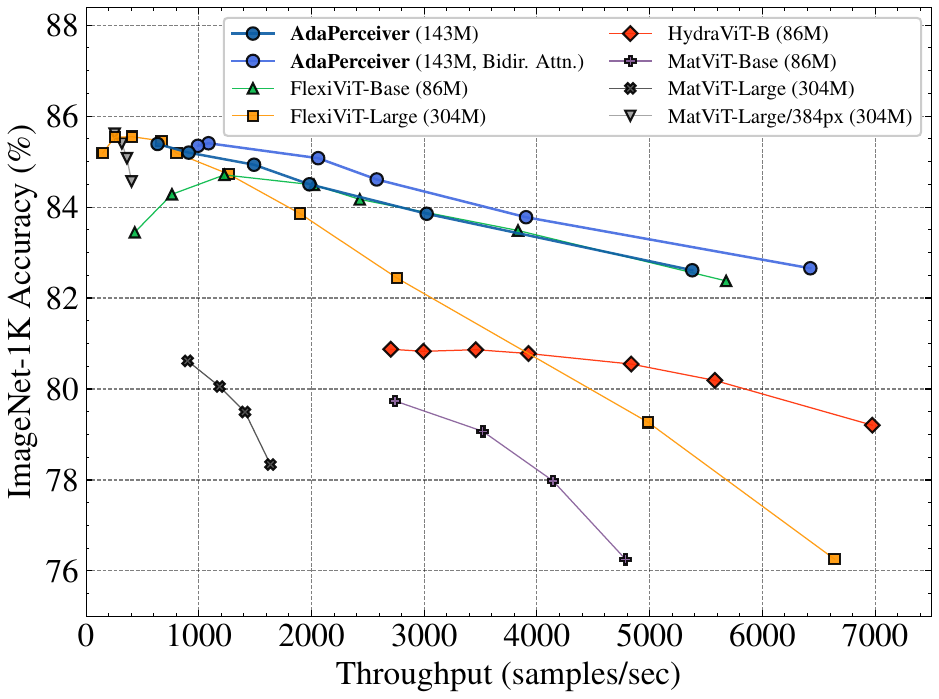}
        \caption{
        Truncated.
        }
        \label{fig:appendix:image-class:pareto-bidir}
    \end{subfigure}
    \hfill
    \begin{subfigure}[t]{0.48\linewidth}
        \centering
        \includegraphics[width=\linewidth]{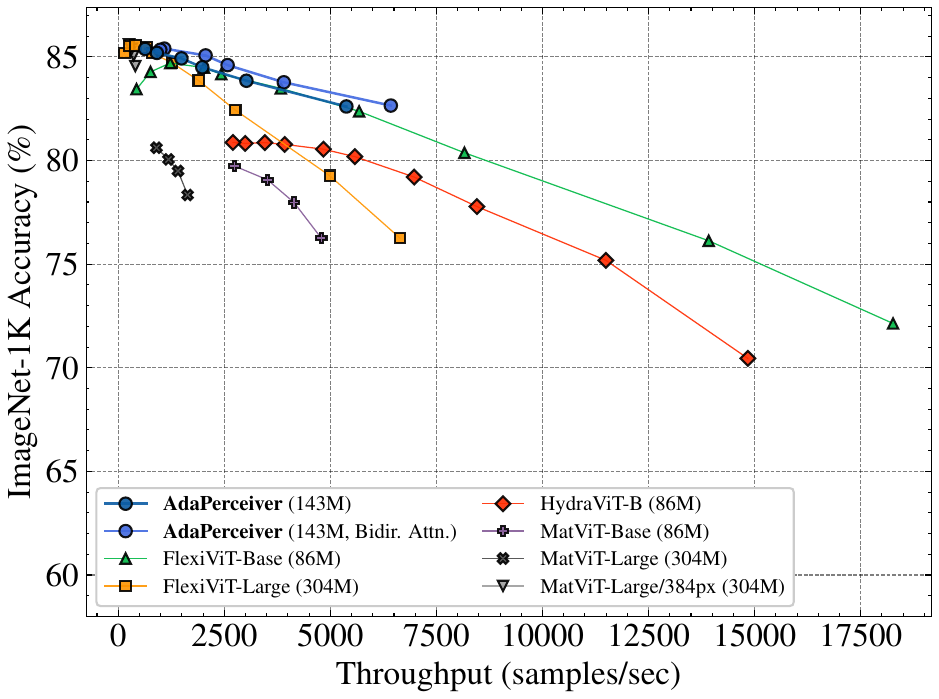}
        \caption{
        Expanded.
        }
        \label{fig:appendix:image-class:pareto-bidir-full}
    \end{subfigure}
    \caption{
    \textbf{ImageNet-1K Evaluation.} 
    Comparison of AdaPerceiver and state-of-the-art adaptive architectures, showing ImageNet-1K accuracy versus throughput.
    \cref{fig:appendix:image-class:pareto-bidir} is identical to \cref{fig:eval:image-class:pareto} but with the addition of bi-directional attention data; 
    \cref{fig:appendix:image-class:pareto-bidir-full} is an expanded version of \cref{fig:appendix:image-class:pareto-bidir}.
    \textit{NB:} Throughput differences between the standard AdaPerceiver its bi-directional form are attributable to changes in the underlying attention implementation.
    AdaPerceiver uses FlexAttention~\cite{dong2024flexattentionprogrammingmodel}, whereas AdaPerceiver (Bidir.) uses PyTorch's \texttt{scaled\_dot\_product\_attention}. 
    }
    \label{fig:appendix:image-class:pareto-bidir-merged}
\end{figure*}

\begin{figure}[ht]
    \centering
    \includegraphics[width=1\linewidth]{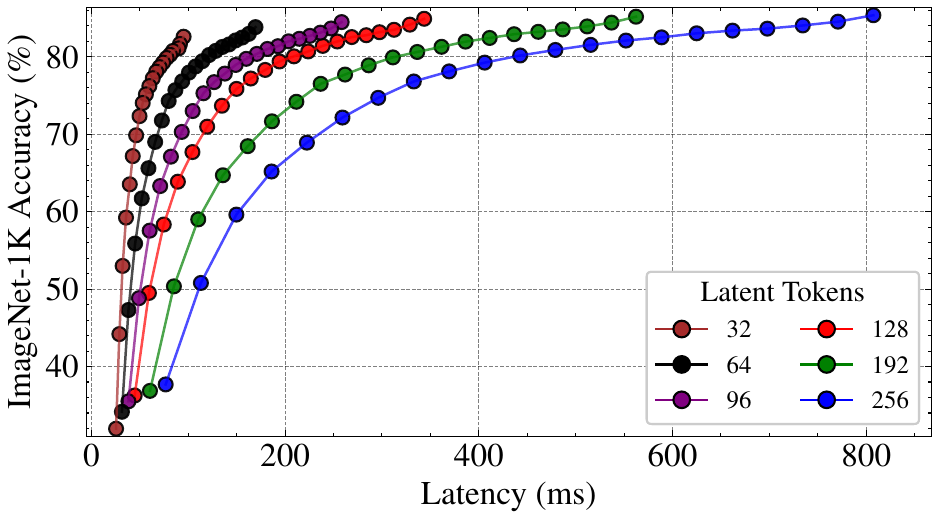}
    \caption{
    \textbf{ImageNet-1K Depth-Token Configuration Tradeoffs.} 
    Accuracy vs. Latency (ms) for AdaPerceiver with varying depths and numbers of latent tokens.
    Importantly, each configuration (point) \textit{does not} require retraining.
    Depth improves accuracy monotonically while increasing latency monotonically.
    Reducing the number of latent tokens substantially decreases latency with minimal accuracy loss.
    }
    \label{fig:appendix:image-class:token-depth}
\end{figure}

We include additional image classification results.
\cref{tab:vit-benchmark} includes an extended comparison with both adaptive architectures and various ViT models. 
\cref{fig:appendix:image-class:pareto-bidir-merged} illustrates the data in \cref{fig:eval:image-class:pareto} with alongside the bi-directional attention variant of AdaPerciever. 
We note that the improvements in throughput with the bi-directional variant are likely due to differences in attention implementation (FlexAttention vs. \texttt{scaled\_dot\_product\_attention}).
\cref{fig:appendix:image-class:token-depth} illustrates how scaling depth and tokens offers different trade-offs; particularly that scaling tokens has minor accuracy degradation for significant latency benefits.

\section{Dense Prediction Results}
\label{sec:appendix:dense}
We present additional results from \cref{sec:eval:dense}.
\cref{sec:appendix:seg} contains additional segmentation results and 
\cref{sec:appendix:depth} contains additional depth estimation results.

\subsection{Semantic Segmentation}
\label{sec:appendix:seg}
\begin{figure}
    \centering
    \includegraphics[width=1.0\linewidth]{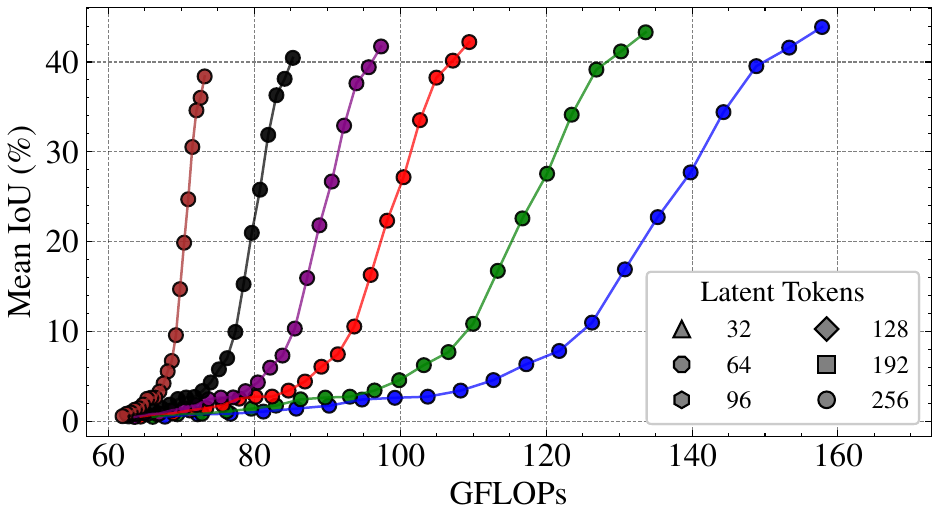}
    \caption{
    \textbf{ADE20K Depth-Token Configuration Tradeoffs.} 
    mIoU vs. GFLOPs for AdaPerceiver with varying depths and numbers of latent tokens.
    }
    \label{fig:appendix:seg:token-depth}
\end{figure}

We include extended experiments on semantic segmentation.
\cref{fig:appendix:seg:token-depth} shows the relationship between GFLOPs and mIoU across depth and tokens, increasing depth improve mIoU monotonically with increasing computational costs. 
\cref{fig:appendix:seg:inter-extrap} illustrates how extrapolating beyond training-time configurations affects mIoU and GFLOPs; extrapolation degrades mIoU, however interpolation retains performance between training points.

\subsection{Depth Estimation}
\label{sec:appendix:depth}
\begin{figure}
    \centering
    \includegraphics[width=1.0\linewidth]{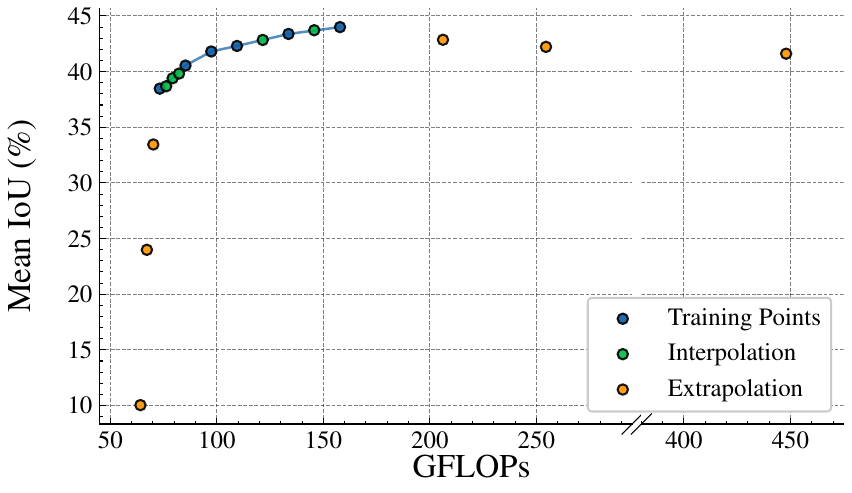}
    \caption{
    \textbf{Effect of Latent Token Interpolation and Extrapolation on ADE20K semantic segmentation.} 
    Similar to \cref{fig:appendix:image-class:inter-extrap}, AdaPerceiver is able to interpolate (\textcolor{ForestGreen}{green} points) between its training token granularities and to extrapolate (\textcolor{Dandelion!90!red}{yellow} points) beyond them.
    Performance degradation appears when extrapolating outside the trained range, with the largest drop occurring when extrapolating below the smallest trained token granularity.
    \textit{N.B. The x-axis contains a break to ease visualization.}
    }
    \label{fig:appendix:seg:inter-extrap}
\end{figure}

\begin{figure}
    \centering
    \includegraphics[width=1.0\linewidth]{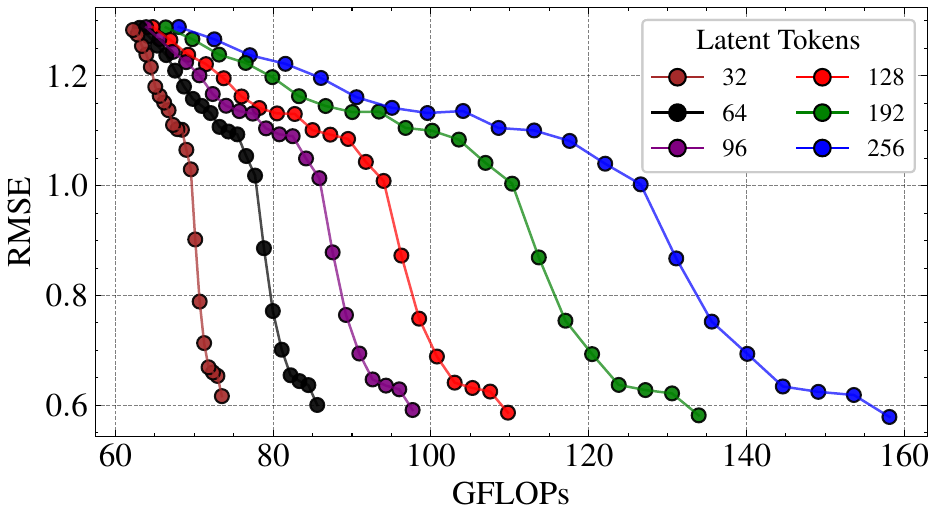}
    \caption{
    \textbf{NYUv2 Depth-Token Configuration Tradeoffs.} 
    RMSE vs. GFLOPs for AdaPerceiver with  varying depths and numbers of latent tokens.
    }
    \label{fig:appendix:depth-est:token-depth}
\end{figure}

We include extended experiments on depth estimation.
\cref{fig:appendix:depth-est:token-depth} shows the relationship between GFLOPs and RMSE across depth and tokens, increasing depth improve mIoU monotonically with increasing computational costs.
\cref{fig:appendix:seg:inter-extrap} illustrates how extrapolating beyond training-time configurations affects RMSE and GFLOPs; extrapolation degrades RMSE, however interpolation retains performance between training points.

\section{Feature Visualizations}
\label{sec:appendix:feat-viz}
\begin{figure}[htbp]
    \centering
    \begin{subfigure}[b]{1.0\linewidth}
        \centering
        \includegraphics[width=\linewidth]{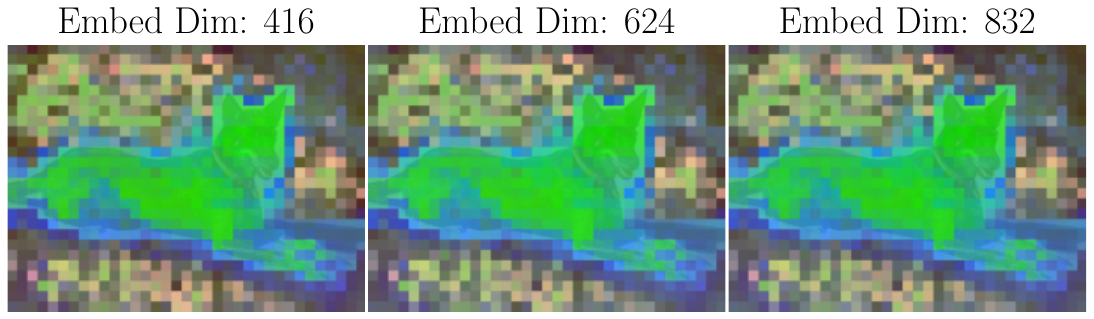}
    \end{subfigure}
    \vfill
    \begin{subfigure}[b]{1.0\linewidth}
        \centering
        \includegraphics[width=\linewidth]{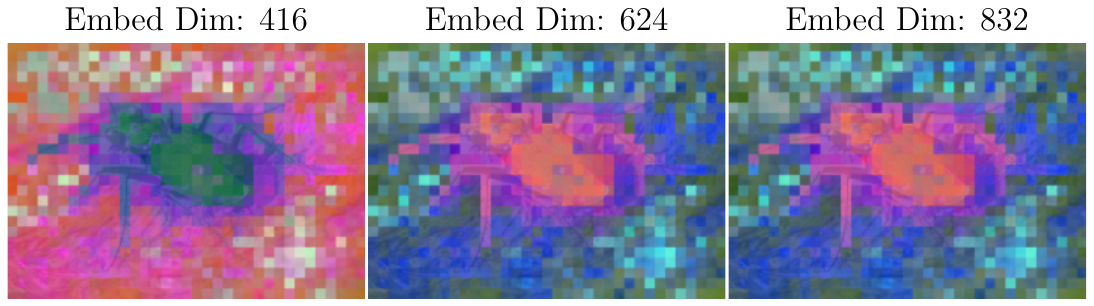}
        \label{fig:sub2}
    \end{subfigure}
    \caption{First three principal components of the patch features from AdaPerceiver when varying the number of embedding tokens (the tokens and depth fixed to their respective maximums). 
    In the top sample, the principal components remain consistent across embedding dimension.
    In the bottom sample, the principal components from 416 $\rightarrow$ 624 width.
    }
    \label{fig:appendix:feat-viz:embed_vary}
\end{figure}

\begin{figure*}[t]
    \centering
    \begin{subfigure}[b]{1.0\linewidth}
        \centering
        \includegraphics[width=\linewidth]{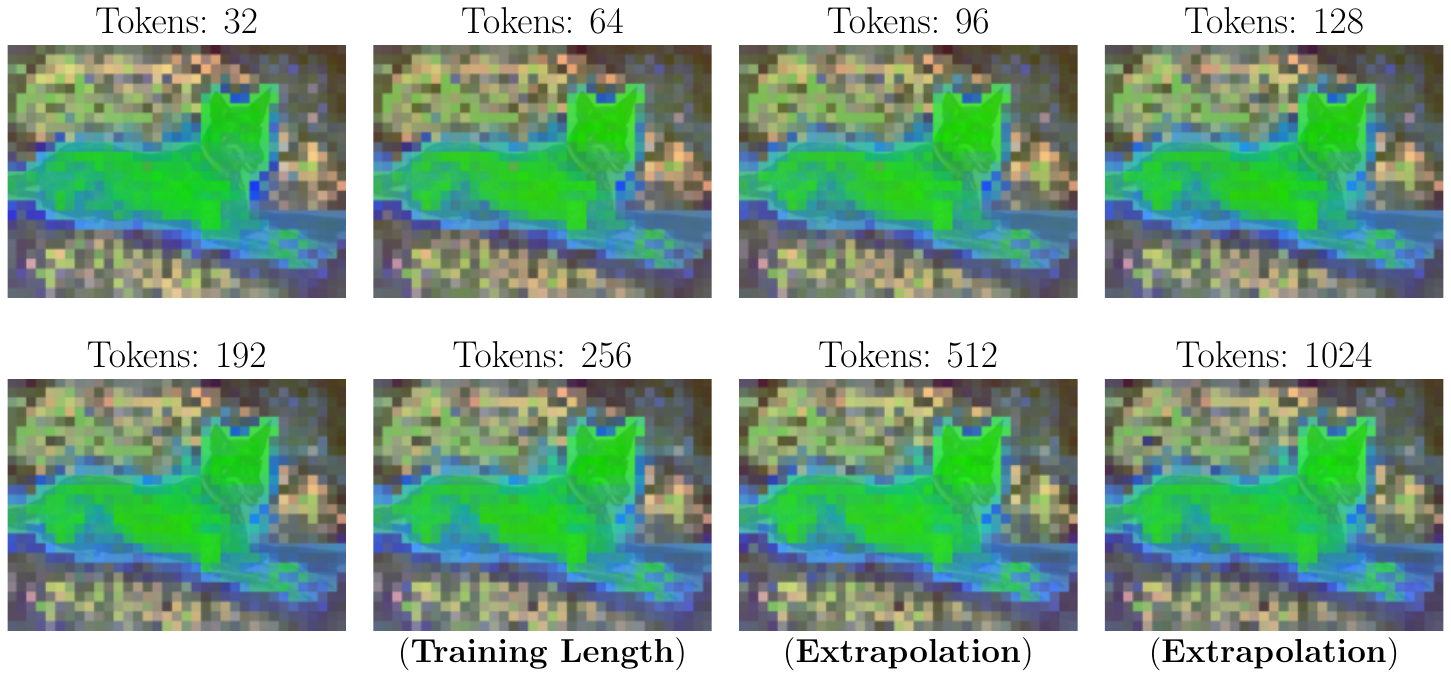}
        \label{fig:appendix:feat-viz:token_vary:1}
    \end{subfigure}
    \vfill
    \begin{subfigure}[b]{1.0\linewidth}
        \centering
        \includegraphics[width=\linewidth]{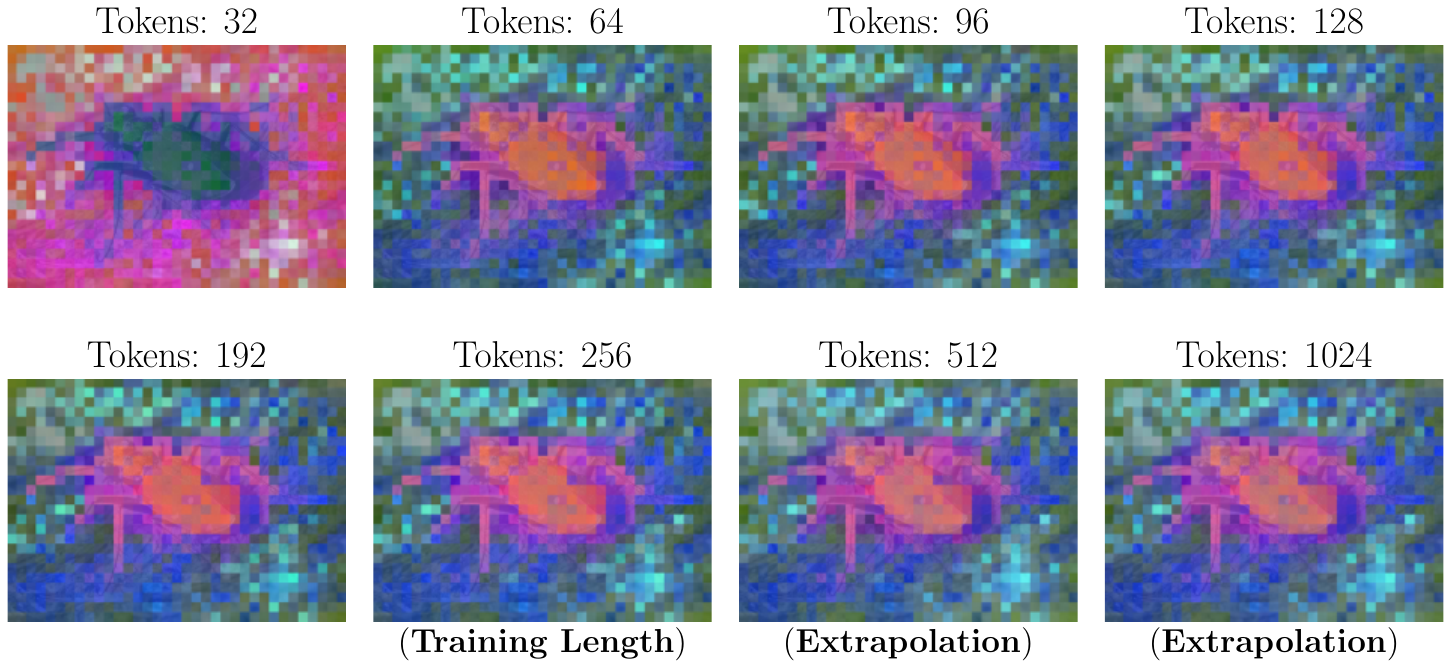}
        \label{fig:appendix:feat-viz:token_vary:2}
    \end{subfigure}
    \caption{
    First three principal components of the patch features from AdaPerceiver when varying the number of latent tokens (the embedding dimension and depth fixed to their respective maximums). 
    In the top sample, the principal components remain consistent across token counts (32 $\rightarrow$ 1024), indicating increasing the number of latent tokens does not change feature maps.
    In the bottom sample, the principal components shift initially from 32 $\rightarrow$ 64 tokens, after which they consistent up to 1024 tokens, suggesting that the model utilizes the additional capacity provided when shifting from 32 to 64 tokens, after which the representations converge.
    In both cases, the principal components \textbf{remain stable when extrapolating past the training length}. 
    }
    \label{fig:appendix:feat-viz:token_vary}
\end{figure*}

\begin{figure*}
    \centering
    \begin{subfigure}[b]{0.465\linewidth}
        \centering
        \includegraphics[width=\linewidth]{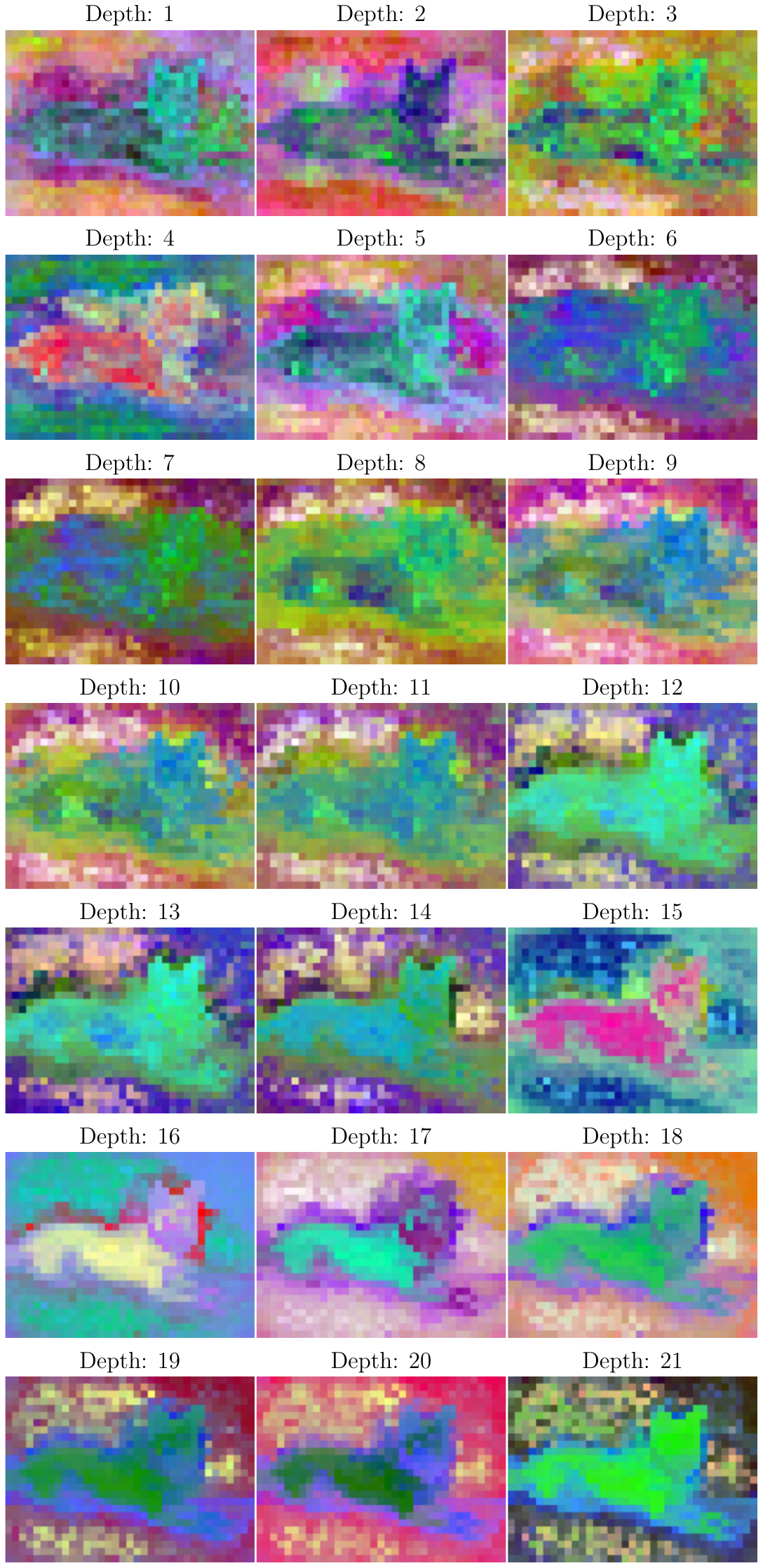}
        \label{fig:appendix:feat-viz:depth_vary:1}
    \end{subfigure}
    \hfill
    \begin{subfigure}[b]{0.47\linewidth}
        \centering
        \includegraphics[width=\linewidth]{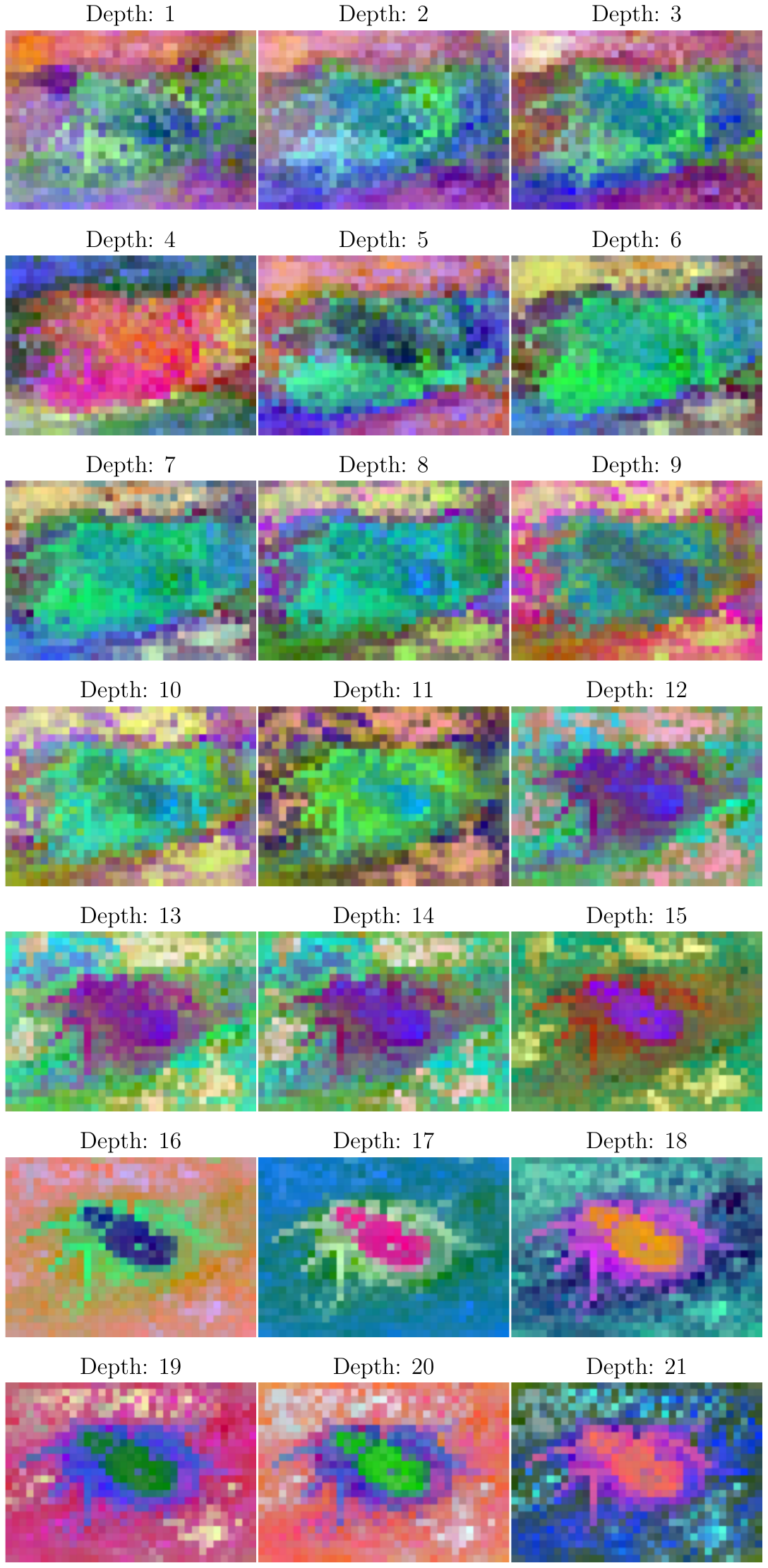}
        \label{fig:appendix:feat-viz:depth_vary:2}
    \end{subfigure}
    
    \caption{First three principal components of patch features across network depth (1–21) in AdaPerceiver.
    In both samples, discernible semantic features emerge with greater depth.
    However, the earliest layer at which discernible features emerge differ with sample. 
    }
    \label{fig:appendix:feat-viz:depth_vary}
\end{figure*}

We include principal component analyses (PCA) of AdaPerceiver's patch features.
\cref{fig:appendix:feat-viz:embed_vary} visualizes principal components as embedding dimension is modulated.
\cref{fig:appendix:feat-viz:token_vary} visualizes principal components as the number of tokens is changed, even extrapolated beyond training length.
\cref{fig:appendix:feat-viz:depth_vary} depicts how depth affects principal components, showing that semantic features emerge at different depths, depending on the input image.

\section{Policies for Adaptivity}
\label{sec:appendix:policies}
Recall, that AdaPerceiver exposes a large space of valid configurations across tokens, depth, and width but does not prescribe which configuration should be used. 
We study the effect of different policies in \cref{sec:eval:policies}, we elaborate on those policies here.

\subsection{Baseline Policy}
To understand the effect of a using a fixed configuration for all inputs, we study a ``Baseline" policy.
This policy is \emph{input-independent}.
We select a fixed configuration (number of tokens $t$, width $w$ and depth $l$) for all inputs.

\subsection{Early Exit (EE) Policy}
To understand the effect of using a simple adaptivity method (early-exiting) in conjunction with a fixed configuration, we study an early-exit policy.
This policy augments the baseline policy; rather than selecting a specific depth, an early-exit threshold is selected.
The early-exit threshold, $\tau$, is the threshold which the confidence of a prediction must exceed to exit early~\cite{jiang2024tracing}. 
During the forward pass the latent tokens are read out, and if the prediction confidence exceeds $\tau$, we exit.
We are able to implement this \textit{without} any further training of our model.

\subsection{RL Policy}
We train a lightweight policy network using REINFORCE~\cite{williams1992simple} to select a token count for an input.
Our policy network definition is shown in \cref{alg:appendix:policy-network}, it consists of MLP-Mixer Block~\cite{tolstikhin2021mlp} and operates on the outputs of the Patch Embedding layer of AdaPerceiver, \ie the input tokens.

We define a discrete action space over $\mathcal{T}$, the token granularities, and our goal is to learn a policy $\pi(t \mid x)$, that associates a token granularity with a given input.
For our reward we use the negative cross-entropy as our reward a with computational cost term:
\begin{equation}
    R(y, \hat{y}, t) = -\text{CrossEntropy}(\hat{y}, y) - \lambda \text{Cost}(t).
\end{equation}
Where, $y$ is the ground-truth label, $\hat{y}$ is the predicted label, and $\lambda$ controls the trade-off between accuracy and computational cost. 
Rather than directly measuring computational, we use a proxy, since computation cost increases monotonically with token count, we increase cost linear with index, \eg if $\mathcal{T}=\{4, 8\}$, then $4$ would have cost $0$ and $8$ would have cost $1$.
Finally, to reduce the variance of REINFORCE, we use the EMA of previous rewards as a baseline. 

\begin{algorithm}[t]
\caption{Policy Network}
\label{alg:appendix:policy-network}
\lstset{style=mocov3, numbers=left, numberstyle=\tiny, numbersep=5pt}
\begin{lstlisting}[language=python,escapechar=@]
class PolicyNetwork(nn.Module):
    def __init__(self, dim, seq_len, token_choices):
        super(PolicyNetwork, self).__init__()
        self.dim = dim
        self.token_choices = token_choices

        self.mixer_block = MixerBlock(dim=dim, seq_len=seq_len)
        self.mixer_block_2 = MixerBlock(dim=dim, seq_len=seq_len)

        # Small fusion MLP after pooling
        self.fuse = nn.Sequential(
            nn.LayerNorm(dim),
            nn.Linear(dim, dim),
            nn.GELU(),
            nn.Linear(dim, dim),
            nn.GELU(),
        )

        self.head_tokens = nn.Linear(dim, len(token_choices), bias=False)

    def forward(self, x):
        x = self.mixer_block(x)
        x = self.mixer_block_2(x)
        x = x.mean(dim=1)

        h = self.fuse(x)

        logits_tokens = self.head_tokens(h)
        return logits_tokens
\end{lstlisting}
\end{algorithm}

\subsection{Optimal Policy}
To characterize the theoretical upper bound on performance we define an oracle ``optimal'' policy.  
Given a trained model, this policy chooses, for each input, the configuration with the least compute that still yields a correct classification.
We perform a grid-search across configurations for each input on the ImageNet-1K validation split, which gives us oracle-like behaviour.
During this search, we record the \emph{minimal compute} configuration that will yield a correct classification --- when running the policy we look-up the minimal configuration for the given input.
This serves as an oracle on ImageNet-1K to help characterize the theoretical upper-bound performance our trained AdaPerceiver model can achieve on this task.

\FloatBarrier
\begin{table*}[ht]
\centering
\caption{
\textbf{ImageNet-1K Cross-Model Evaluation.}
Comparison of Vision Transformer (ViT) variants on ImageNet-1K. 
Metrics include classification accuracy, inference latency (mean per forward pass), and computational cost in GFLOPs for both forward and backward passes. 
Latency measured at batch size of 512.
}
\label{tab:vit-benchmark}
\begin{tabular}{lrrrrrr}
\toprule
\textbf{Model} & \textbf{Params (M)} & \textbf{Accuracy (\%)} & \textbf{Latency (ms)} & \textbf{Fwd GFLOPs} & \textbf{Bwd GFLOPs} \\
\midrule
DeiT-Ti/16 & 5.7 & 68.96 & 29.4 & 6.4 & 12.7 \\
DeiT-S/16 & 22.1 & 78.21 & 66.3 & 25.3 & 50.4 \\
DeiT-B/16 & 86.6 & 80.79 & 177.1 & 100.9 & 201.3 \\
DeiT3-S/16 & 22.1 & 80.83 & 66.2 & 25.3 & 50.4 \\
DeiT3-B/16 & 86.6 & 83.22 & 176.2 & 100.9 & 201.3 \\
DeiT3-L/16 & 304.4 & 84.23 & 545.5 & 357.6 & 714.5 \\
ViT-H/14 & 632.0 & 87.11 & 1504.8 & 970.9 & 1941.1 \\
SoViT-150M/16 & 136.1 & 87.27 & 403.2 & 207.6 & 414.5 \\
\midrule
MatViT-B ($w=96$)  & 86.6 & 76.3 & 107.0 & 49.4 & 98.2 \\
MatViT-B ($w=192$) & 86.6 & 77.9 & 123.6 & 59.1 & 117.7 \\
MatViT-B ($w=384$) & 86.6 & 79.1 & 145.5 & 78.7 & 156.8 \\
MatViT-B ($w=768$) & 86.6 & 79.7 & 186.7 & 117.7 & 234.8 \\
\midrule
MatViT-L ($w=128$)  & 304.3 & 78.3 & 313.0 & 174.3 & 347.8 \\
MatViT-L ($w=256$)  & 304.3 & 79.5 & 363.1 & 209.0 & 417.2 \\
MatViT-L ($w=512$)  & 304.3 & 80.1 & 432.1 & 278.4 & 556.0 \\
MatViT-L ($w=1024$) & 304.3 & 80.6 & 567.9 & 417.2 & 833.7 \\
\midrule
MatViT-L/384px ($w=128$)  & 304.7 & 84.6 & 1268.4 & 510.6 & 1018.7 \\
MatViT-L/384px ($w=256$)  & 304.7 & 85.1 & 1411.2 & 612.1 & 1222.0 \\
MatViT-L/384px ($w=512$)  & 304.7 & 85.4 & 1611.5 & 815.4 & 1628.6 \\
MatViT-L/384px ($w=1024$) & 304.7 & 85.6 & 2013.4 & 1222.0 & 2441.8 \\
\midrule
HydraViT ($w=192$) & 86.6 & 70.5 & 34.7 & 4.27 & 8.45 \\
HydraViT ($w=256$) & 86.6 & 75.2 & 44.8 & 7.55 & 14.99 \\
HydraViT ($w=320$) & 86.6 & 77.8 & 60.8 & 11.76 & 23.38 \\
HydraViT ($w=384$) & 86.6 & 79.2 & 73.9 & 16.91 & 33.64 \\
HydraViT ($w=448$) & 86.6 & 80.2 & 92.0 & 22.98 & 45.75 \\
HydraViT ($w=512$) & 86.6 & 80.5 & 106.3 & 29.98 & 59.73 \\
HydraViT ($w=576$) & 86.6 & 80.8 & 131.0 & 37.91 & 75.56 \\
HydraViT ($w=640$) & 86.6 & 80.9 & 148.4 & 46.77 & 93.25 \\
HydraViT ($w=704$) & 86.6 & 80.8 & 171.7 & 56.56 & 112.80 \\
HydraViT ($w=768$) & 86.6 & 80.9 & 190.3 & 67.28 & 134.21 \\
\bottomrule
\end{tabular}
\end{table*}
\FloatBarrier

\begin{table*}[ht]
\centering
\ContinuedFloat
\caption[]{\textbf{(Continued) ImageNet-1K Cross-Model Evaluation.}}
\begin{tabular}{lrrrrrr}
\toprule
\textbf{Model} & \textbf{Params (M)} & \textbf{Accuracy (\%)} & \textbf{Latency (ms)} & \textbf{Fwd GFLOPs} & \textbf{Bwd GFLOPs} \\
\midrule
FlexViT-B  ($ps=48$) & 91.2 & 72.1 & 28.0   & 13.8  & 27.0 \\
FlexViT-B  ($ps=40$) & 89.5 & 76.1 & 36.8   & 19.4  & 38.3 \\
FlexViT-B  ($ps=30$) & 87.9 & 80.4 & 62.7   & 33.7  & 66.8 \\
FlexViT-B  ($ps=24$) & 87.2 & 82.4 & 90.2   & 52.0  & 103.5 \\
FlexViT-B  ($ps=20$) & 86.9 & 83.5 & 133.6  & 74.4  & 148.3 \\
FlexViT-B  ($ps=16$) & 86.6 & 84.2 & 210.9  & 115.7 & 230.9 \\
FlexViT-B  ($ps=15$) & 86.5 & 84.5 & 253.2  & 131.5 & 262.5 \\
FlexViT-B  ($ps=12$) & 86.5 & 84.7 & 417.3  & 204.9 & 409.2 \\
FlexViT-B  ($ps=10$) & 86.5 & 84.3 & 672.1  & 294.6 & 588.6 \\
FlexViT-B ($ps=8$)  & 86.7 & 83.4 & 1185.2 & 459.7 & 918.9 \\
\midrule
FlexiViT-L ($ps=48$) & 310.4 & 76.3 & 77.1 & 47.8 & 94.9 \\
FlexiViT-L ($ps=40$) & 308.3 & 79.3 & 103  & 67.8 & 134 \\
FlexiViT-L ($ps=30$) & 306.2 & 82.4 & 185  & 118  & 236 \\
FlexiViT-L ($ps=24$) & 305.2 & 83.9 & 270  & 184  & 367 \\
FlexiViT-L ($ps=20$) & 304.7 & 84.7 & 404  & 263  & 526 \\
FlexiViT-L ($ps=16$) & 304.4 & 85.2 & 638  & 410  & 819 \\
FlexiViT-L ($ps=15$) & 304.4 & 85.4 & 765  & 466  & 932 \\
FlexiViT-L ($ps=12$) & 304.1 & 85.5 & 1257 & 727  & 1453 \\
FlexiViT-L ($ps=10$) & 304.2 & 85.5 & 2001 & 1046 & 2091 \\
FlexiViT-L ($ps=8$)  & 304.4 & 85.2 & 3469 & 1633 & 3266 \\
\midrule
AdaPerceiver ($t=32$) &  143.8 & 82.6 &  95.2 & 16.2 & 44.1 \\
AdaPerceiver ($t=64$) &  143.8 & 83.9 & 169.4 & 28.3 & 76.8 \\
AdaPerceiver ($t=96$) &  143.8 & 84.5 & 258.4 & 40.4 & 109.6 \\
AdaPerceiver ($t=128$) & 143.8 & 84.9 & 343.6 & 52.5 & 142.4  \\
AdaPerceiver ($t=192$) & 143.8 & 85.2 & 562.3 & 88.7 & 240.7 \\
AdaPerceiver ($t=256$) & 143.8 & 85.4 & 807.4 & 100.8 & 273.5 \\
\midrule
AdaPerceiver, Bidir. ($t=32$) &  143.8 & 82.6 & 79.7 & 16.2 & 44.1 \\
AdaPerceiver, Bidir. ($t=64$) &  143.8 & 83.7 & 131.2 & 28.3 & 76.8 \\
AdaPerceiver, Bidir. ($t=96$) &  143.8 & 84.6 & 198.6 & 40.4 & 109.6 \\
AdaPerceiver, Bidir. ($t=128$) & 143.8 & 85.0 & 248.6 & 52.5 & 142.4  \\
AdaPerceiver, Bidir. ($t=192$) & 143.8 & 85.4 & 471.3 & 88.7 & 240.7 \\
AdaPerceiver, Bidir. ($t=256$) & 143.8 & 85.3 & 516.0 & 100.8 & 273.5 \\
\bottomrule
\end{tabular}
\end{table*}
\FloatBarrier

\begin{table}[!tbh]
\centering
\caption{
\textbf{Adaptivity Policy Evaluation.}
Accuracy and computational cost (GFLOPs) for configuration selection policies applied to AdaPerceiver on image classification.
Utilizing early-exiting often acts as a ``free-lunch" allowing for the reduction in compute costs with little to no degradation in accuracy.
\textit{N.B.} The ``Optimal" policy is only theoretical and impractical to realize.
}
\setlength{\tabcolsep}{5.2pt}
\begin{tabular}{lrr}
\toprule
\textbf{Policy} & \textbf{Accuracy} (\%) \blueuparrow & \textbf{GFLOPs} \bluedownarrow \\
\midrule
Baseline ($t=32$)  & 82.7 & 16.2 \\
Baseline ($t=64$)  & 83.8 & 28.3 \\
Baseline ($t=96$) & 84.5 & 40.4 \\
Baseline ($t=128$) & 85.0 & 52.5 \\
Baseline ($t=192$) & 85.3 & 76.7 \\
Baseline ($t=256$) & 85.4 & 100.8 \\
\midrule
EE ($t=32, \;\; \tau=0.90$) & 82.4 & 12.5 \\
EE ($t=64, \;\; \tau=0.90$) & 83.6 & 19.9 \\
EE ($t=128, \tau=0.90$) & 84.7 & 35.0 \\
EE ($t=192, \tau=0.90$) & 85.1 & 51.2 \\
EE ($t=256, \tau=0.90$) & 85.3 & 66.8 \\
EE ($t=256, \tau=0.95$) & 85.4 &  76.5 \\
\midrule
RL (tokens only) & 83.9 & 32.0 \\
RL (tokens, $\tau=0.9$) & 85.0 & 46.9  \\
\midrule
Optimal & 93.6 & 32.5 \\
\bottomrule
\end{tabular}
\label{tab:results:policy-full}
\end{table}

\end{document}